\documentclass[11pt]{article}

\usepackage[utf8]{inputenc}
\usepackage[T1]{fontenc}
\usepackage[margin=1in]{geometry}
\usepackage{booktabs}
\usepackage{graphicx}
\usepackage[skip=6pt]{caption}
\usepackage{placeins}
\usepackage{needspace}
\usepackage{xcolor}
\usepackage[hidelinks,
  pdftitle={PlotPick: AI-powered batch extraction of numerical data from scientific figures},
  pdfauthor={Tommy Carstensen}]{hyperref}
\usepackage{xurl}
\usepackage{orcidlink}
\usepackage{authblk}

\title{PlotPick: AI-powered batch extraction of numerical data from scientific figures}

\author[1]{Tommy Carstensen\,\orcidlink{0000-0002-3672-9931}}
\affil[1]{Copenhagen Research Centre for Biological and Precision Psychiatry,
  Mental Health Centre Copenhagen, Copenhagen University Hospital, Copenhagen, Denmark}

\date{}

\newcommand{\Tolerance}{5}
\newcommand{\ToleranceExample}{20.0}
\newcommand{\ToleranceExampleLow}{19.0}
\newcommand{\ToleranceExampleHigh}{21.0}
\newcommand{\SmdExampleMeans}{20.0 and 22.0}
\newcommand{\SmdExampleSD}{5}
\newcommand{\SmdExampleTrue}{0.40}
\newcommand{\SmdExampleLow}{$-$0.02}
\newcommand{\SmdExampleHigh}{0.82}
\newcommand{\ImageMaxSide}{1{,}024}
\newcommand{\BootstrapDraws}{10{,}000}
\newcommand{\BorderlinePoints}{0.1}
\newcommand{\HeatmapFloor}{50}
\newcommand{\ProviderCount}{four}
\newcommand{\MistralModelCount}{three}
\newcommand{\NonMistralModelCount}{six}
\newcommand{\ChartXTypesTotal}{18}
\newcommand{\ChartXValItems}{4{,}848}
\newcommand{\ChartXTestItems}{1{,}152}
\newcommand{\ChartXTypesOther}{12}
\newcommand{\ChartXTypeCount}{six}
\newcommand{\ChartXNonBoxTypeCount}{five}
\newcommand{\ChartXPerType}{50}
\newcommand{\DePlotMaxNewTokens}{512}
\newcommand{\ChartXDePlotLongReplies}{two}
\newcommand{\ChartXCorrections}{eight}

\newcommand{\ChartXFoneGeminiFlash}{96.0}
\newcommand{\ChartXFoneCIGeminiFlash}{\mbox{[94.8, 97.2]}}
\newcommand{\ChartXRecGeminiFlash}{95.8}
\newcommand{\ChartXNGeminiFlash}{300}
\newcommand{\ChartXDeltaGeminiFlash}{+21.7}
\newcommand{\ChartXDeltaCIGeminiFlash}{\mbox{[+18.6, +25.0]}}
\newcommand{\ChartXFoneGeminiFlashLite}{95.1}
\newcommand{\ChartXFoneCIGeminiFlashLite}{\mbox{[93.8, 96.4]}}
\newcommand{\ChartXRecGeminiFlashLite}{94.3}
\newcommand{\ChartXNGeminiFlashLite}{300}
\newcommand{\ChartXDeltaGeminiFlashLite}{+20.8}
\newcommand{\ChartXDeltaCIGeminiFlashLite}{\mbox{[+17.8, +24.0]}}
\newcommand{\ChartXFoneGptMini}{93.4}
\newcommand{\ChartXFoneCIGptMini}{\mbox{[91.8, 94.9]}}
\newcommand{\ChartXRecGptMini}{93.4}
\newcommand{\ChartXNGptMini}{300}
\newcommand{\ChartXDeltaGptMini}{+19.1}
\newcommand{\ChartXDeltaCIGptMini}{\mbox{[+16.2, +22.1]}}
\newcommand{\ChartXFoneSonnet}{92.2}
\newcommand{\ChartXFoneCISonnet}{\mbox{[90.8, 93.6]}}
\newcommand{\ChartXRecSonnet}{92.0}
\newcommand{\ChartXNSonnet}{300}
\newcommand{\ChartXDeltaSonnet}{+17.9}
\newcommand{\ChartXDeltaCISonnet}{\mbox{[+15.2, +20.8]}}
\newcommand{\ChartXFoneGptNano}{89.0}
\newcommand{\ChartXFoneCIGptNano}{\mbox{[87.0, 90.8]}}
\newcommand{\ChartXRecGptNano}{89.7}
\newcommand{\ChartXNGptNano}{300}
\newcommand{\ChartXDeltaGptNano}{+14.6}
\newcommand{\ChartXDeltaCIGptNano}{\mbox{[+11.9, +17.5]}}
\newcommand{\ChartXFoneHaiku}{88.7}
\newcommand{\ChartXFoneCIHaiku}{\mbox{[86.9, 90.4]}}
\newcommand{\ChartXRecHaiku}{88.6}
\newcommand{\ChartXNHaiku}{299}
\newcommand{\ChartXDeltaHaiku}{+14.1}
\newcommand{\ChartXDeltaCIHaiku}{\mbox{[+11.3, +17.0]}}
\newcommand{\ChartXFoneMistralMedium}{84.8}
\newcommand{\ChartXFoneCIMistralMedium}{\mbox{[82.5, 87.1]}}
\newcommand{\ChartXRecMistralMedium}{84.8}
\newcommand{\ChartXNMistralMedium}{300}
\newcommand{\ChartXDeltaMistralMedium}{+10.5}
\newcommand{\ChartXDeltaCIMistralMedium}{\mbox{[+8.0, +13.0]}}
\newcommand{\ChartXFoneMistralSmall}{80.2}
\newcommand{\ChartXFoneCIMistralSmall}{\mbox{[77.7, 82.6]}}
\newcommand{\ChartXRecMistralSmall}{81.4}
\newcommand{\ChartXNMistralSmall}{300}
\newcommand{\ChartXDeltaMistralSmall}{+5.8}
\newcommand{\ChartXDeltaCIMistralSmall}{\mbox{[+3.1, +8.6]}}
\newcommand{\ChartXFoneMistralLarge}{79.1}
\newcommand{\ChartXFoneCIMistralLarge}{\mbox{[76.5, 81.5]}}
\newcommand{\ChartXRecMistralLarge}{81.1}
\newcommand{\ChartXNMistralLarge}{300}
\newcommand{\ChartXDeltaMistralLarge}{+4.7}
\newcommand{\ChartXDeltaCIMistralLarge}{\mbox{[+1.8, +7.7]}}
\newcommand{\ChartXFoneDePlot}{74.3}
\newcommand{\ChartXFoneCIDePlot}{\mbox{[71.0, 77.6]}}
\newcommand{\ChartXRecDePlot}{74.7}
\newcommand{\ChartXNDePlot}{300}
\newcommand{\ChartXFoneDePlotHaikuItems}{74.6}
\newcommand{\ChartXFoneHaikuZero}{88.4}
\newcommand{\ChartXDeltaHaikuZero}{+14.0}

\newcommand{\ChartXDePlotTitlesWithNumber}{220}
\newcommand{\ChartXFoneDePlotTitled}{72.8}

\newcommand{\ChartXTruthRepeats}{171}
\newcommand{\ChartXFoneDePlotRepeats}{73.6}
\newcommand{\ChartXLabelShareMin}{4}
\newcommand{\ChartXLabelShareMax}{50}
\newcommand{\ChartXLabelFallMax}{1.9}
\newcommand{\ChartXLabelLowestDelta}{+2.8}
\newcommand{\ChartXLabelLowestDeltaCI}{\mbox{[$-$0.2, +6.0]}}
\newcommand{\ChartXHistogramNumeric}{25}

\newcommand{\ChartXFoneVLMMin}{79.1}
\newcommand{\ChartXFoneVLMMax}{96.0}
\newcommand{\ChartXLowestVLMName}{Mistral Large 3}
\newcommand{\ChartXGapLowestVLM}{4.7}
\newcommand{\ChartXGapLowestVLMCI}{\mbox{[+1.8, +7.7]}}
\newcommand{\ChartXGapHighestVLM}{21.7}
\newcommand{\ChartXModelCount}{nine}
\newcommand{\ChartXSweepCount}{seven}
\newcommand{\ChartXTrailingModels}{two}
\newcommand{\ChartXTrailingCells}{six}
\newcommand{\ChartXTotalCells}{54}
\newcommand{\ChartXTrailingList}{Mistral Small 4 on plain bar charts, bar charts with data labels and plain line charts; Mistral Large 3 on plain bar charts, plain line charts and histograms}
\newcommand{\ChartXTrailingSeparable}{Mistral Large 3 on plain line charts}
\newcommand{\ChartXCellsLeadSeparable}{41}
\newcommand{\ChartXCellsTrailSeparable}{1}
\newcommand{\ChartXCellsSpanZero}{12}
\newcommand{\ChartXTrailingCellsRecall}{seven}
\newcommand{\ChartXBoxDePlot}{24.8}
\newcommand{\ChartXBoxVLMMin}{64.2}
\newcommand{\ChartXBoxVLMMax}{97.3}
\newcommand{\ChartXHistogramDePlot}{86.2}
\newcommand{\ChartXMarginBox}{55.0}
\newcommand{\ChartXMarginMeanMin}{5.3}
\newcommand{\ChartXMarginMeanMax}{8.3}
\newcommand{\ChartXMarginLowestMin}{7.0}
\newcommand{\ChartXMarginLowestMax}{1.5}
\newcommand{\ChartXNonBoxLeadMin}{4.7}
\newcommand{\ChartXNonBoxLeadMax}{11.6}
\newcommand{\ChartXNonBoxLevel}{Mistral Small 4 $-$1.2 \mbox{[$-$3.7, +1.3]}}
\newcommand{\ChartXNonBoxBehind}{Mistral Large 3 $-$2.9 \mbox{[$-$5.6, $-$0.4]}}
\newcommand{\ChartXNonBoxBehindTitled}{$-$1.1 \mbox{[$-$3.7, +1.5]}}
\newcommand{\ChartXBoxShareMin}{55}
\newcommand{\ChartXBoxShareMax}{67}
\newcommand{\ChartXTopCount}{four}
\newcommand{\ChartXTopCellMin}{6.1}
\newcommand{\ChartXTopCellMax}{17.8}
\newcommand{\ChartXLabelledSixMin}{89.1}
\newcommand{\ChartXLabelledSixMax}{98.0}
\newcommand{\ChartXUnlabelledTypeCount}{four}
\newcommand{\ChartXHaikuUnlabelledMin}{79.8}
\newcommand{\ChartXHaikuUnlabelledMax}{92.9}
\newcommand{\ChartXBoxDePlotNumbers}{5.2}
\newcommand{\ChartXBoxTruthNumbers}{26.9}
\newcommand{\ChartXBoxDePlotCeiling}{32.6}
\newcommand{\ChartXAdjacentNotSeparable}{GPT-5.4 nano and Claude Haiku 4.5 (+0.4 \mbox{[$-$1.3, +2.0]}); Mistral Small 4 and Mistral Large 3 (+1.1 \mbox{[$-$0.5, +2.7]})}
\newcommand{\ChartXAdjacentBorderline}{GPT-5.4 mini and Claude Sonnet 4.6 (+1.16 \mbox{[+0.03, +2.25]})}
\newcommand{\ChartXLabelGainMin}{0.7}
\newcommand{\ChartXLabelGainMax}{18.5}
\newcommand{\ChartXLabelComparisons}{20}
\newcommand{\ChartXLabelSpanZero}{11}
\newcommand{\DevBoxFoneMin}{80}
\newcommand{\DevBoxFoneMax}{98}
\newcommand{\ValBoxSixFoneMin}{76}
\newcommand{\ValBoxSixFoneMax}{97}
\newcommand{\DevPerTypeMin}{50}
\newcommand{\DevPerTypeMax}{100}
\newcommand{\ChartXDevModels}{six}
\newcommand{\ChartXDevGapMin}{0.8}
\newcommand{\ChartXDevGapMax}{3.4}
\newcommand{\DevOtherN}{120}
\newcommand{\DevOtherDePlot}{36.5}
\newcommand{\DevSixN}{60}
\newcommand{\DevDePlotPerType}{10}
\newcommand{\DevDePlotN}{180}
\newcommand{\DevOtherDePlotLongReplies}{25}
\newcommand{\DevSixFoneDePlot}{73.5}
\newcommand{\DevOtherFoneDePlot}{36.5}
\newcommand{\DevSixFoneHaiku}{91.2}
\newcommand{\DevOtherFoneHaiku}{67.4}
\newcommand{\DevSixFoneSonnet}{93.9}
\newcommand{\DevOtherFoneSonnet}{71.8}
\newcommand{\DevOtherNHaiku}{119}
\newcommand{\DevOtherDeltaHaiku}{+33.3}
\newcommand{\DevOtherDeltaCIHaiku}{\mbox{[+28.6, +37.8]}}
\newcommand{\DevSixDeltaHaiku}{+16.0}
\newcommand{\DevSixDeltaCIHaiku}{\mbox{[+9.4, +22.9]}}
\newcommand{\DevOtherNSonnet}{120}
\newcommand{\DevOtherDeltaSonnet}{+37.2}
\newcommand{\DevOtherDeltaCISonnet}{\mbox{[+32.9, +41.5]}}
\newcommand{\DevSixDeltaSonnet}{+19.1}
\newcommand{\DevSixDeltaCISonnet}{\mbox{[+12.9, +25.7]}}

\newcommand{\AliasRunCountCap}{Three}
\newcommand{\AliasRunCount}{three}
\newcommand{\AliasDevRunCount}{four}
\newcommand{\AliasLeadCountCap}{Two}
\newcommand{\MinistralN}{295}
\newcommand{\MinistralFone}{76.7}
\newcommand{\MinistralDePlotFone}{73.9}
\newcommand{\MinistralDelta}{+2.8}
\newcommand{\MinistralDeltaCI}{\mbox{[$-$0.5, +6.1]}}
\newcommand{\MinistralRecDelta}{+3.2}
\newcommand{\MinistralRecDeltaCI}{\mbox{[$-$0.2, +6.7]}}
\newcommand{\MinistralTrailingTypes}{four}
\newcommand{\MinistralMissing}{five}
\newcommand{\MinistralFoneZero}{75.5}
\newcommand{\MinistralDeltaZero}{+1.2}
\newcommand{\MinistralDeltaZeroCI}{\mbox{[$-$2.4, +4.6]}}

\newcommand{\PlotQASubsetN}{1{,}000}
\newcommand{\PlotQASubsetHorizontal}{898}
\newcommand{\PlotQASubsetOther}{102}
\newcommand{\PlotQAN}{529}
\newcommand{\PlotQASkipped}{471}
\newcommand{\PlotQAHorizontalN}{427}
\newcommand{\PlotQAOtherN}{102}
\newcommand{\PlotQAModelCount}{six}
\newcommand{\PlotQARepeatedSeries}{two}
\newcommand{\PlotQABestGeminiFlash}{89.2}
\newcommand{\PlotQAHorizontalGeminiFlash}{87.8}
\newcommand{\PlotQAOtherGeminiFlash}{94.7}
\newcommand{\PlotQAWholeGeminiFlash}{37.5}
\newcommand{\PlotQADeltaGeminiFlash}{+2.2}
\newcommand{\PlotQADeltaCIGeminiFlash}{\mbox{[+0.3, +4.1]}}
\newcommand{\PlotQAWholeDeltaGeminiFlash}{+1.0}
\newcommand{\PlotQAWholeDeltaCIGeminiFlash}{\mbox{[+0.2, +1.9]}}
\newcommand{\PlotQAHorizontalDeltaGeminiFlash}{+1.5}
\newcommand{\PlotQAHorizontalDeltaCIGeminiFlash}{\mbox{[$-$0.6, +3.7]}}
\newcommand{\PlotQAOtherDeltaGeminiFlash}{+5.0}
\newcommand{\PlotQAOtherDeltaCIGeminiFlash}{\mbox{[+1.0, +9.6]}}
\newcommand{\PlotQABestSonnet}{89.0}
\newcommand{\PlotQAHorizontalSonnet}{87.8}
\newcommand{\PlotQAOtherSonnet}{94.1}
\newcommand{\PlotQAWholeSonnet}{37.3}
\newcommand{\PlotQADeltaSonnet}{+2.1}
\newcommand{\PlotQADeltaCISonnet}{\mbox{[$-$0.06, +4.2]}}
\newcommand{\PlotQAWholeDeltaSonnet}{+0.8}
\newcommand{\PlotQAWholeDeltaCISonnet}{\mbox{[$-$0.1, +1.8]}}
\newcommand{\PlotQAHorizontalDeltaSonnet}{+1.5}
\newcommand{\PlotQAHorizontalDeltaCISonnet}{\mbox{[$-$0.9, +3.9]}}
\newcommand{\PlotQAOtherDeltaSonnet}{+4.5}
\newcommand{\PlotQAOtherDeltaCISonnet}{\mbox{[$-$0.2, +9.4]}}
\newcommand{\PlotQABestGeminiFlashLite}{83.2}
\newcommand{\PlotQAHorizontalGeminiFlashLite}{80.9}
\newcommand{\PlotQAOtherGeminiFlashLite}{93.1}
\newcommand{\PlotQAWholeGeminiFlashLite}{35.9}
\newcommand{\PlotQADeltaGeminiFlashLite}{$-$3.8}
\newcommand{\PlotQADeltaCIGeminiFlashLite}{\mbox{[$-$5.9, $-$1.5]}}
\newcommand{\PlotQAWholeDeltaGeminiFlashLite}{$-$0.6}
\newcommand{\PlotQAWholeDeltaCIGeminiFlashLite}{\mbox{[$-$1.6, +0.5]}}
\newcommand{\PlotQAHorizontalDeltaGeminiFlashLite}{$-$5.5}
\newcommand{\PlotQAHorizontalDeltaCIGeminiFlashLite}{\mbox{[$-$8.1, $-$2.9]}}
\newcommand{\PlotQAOtherDeltaGeminiFlashLite}{+3.5}
\newcommand{\PlotQAOtherDeltaCIGeminiFlashLite}{\mbox{[$-$0.03, +7.5]}}
\newcommand{\PlotQABestGptMini}{74.0}
\newcommand{\PlotQAHorizontalGptMini}{70.3}
\newcommand{\PlotQAOtherGptMini}{89.8}
\newcommand{\PlotQAWholeGptMini}{32.3}
\newcommand{\PlotQADeltaGptMini}{$-$13.0}
\newcommand{\PlotQADeltaCIGptMini}{\mbox{[$-$15.5, $-$10.4]}}
\newcommand{\PlotQAWholeDeltaGptMini}{$-$4.1}
\newcommand{\PlotQAWholeDeltaCIGptMini}{\mbox{[$-$5.3, $-$3.0]}}
\newcommand{\PlotQAHorizontalDeltaGptMini}{$-$16.1}
\newcommand{\PlotQAHorizontalDeltaCIGptMini}{\mbox{[$-$19.1, $-$13.2]}}
\newcommand{\PlotQAOtherDeltaGptMini}{+0.1}
\newcommand{\PlotQAOtherDeltaCIGptMini}{\mbox{[$-$3.6, +4.2]}}
\newcommand{\PlotQABestHaiku}{70.2}
\newcommand{\PlotQAHorizontalHaiku}{67.1}
\newcommand{\PlotQAOtherHaiku}{82.9}
\newcommand{\PlotQAWholeHaiku}{31.8}
\newcommand{\PlotQADeltaHaiku}{$-$16.8}
\newcommand{\PlotQADeltaCIHaiku}{\mbox{[$-$19.8, $-$13.9]}}
\newcommand{\PlotQAWholeDeltaHaiku}{$-$4.6}
\newcommand{\PlotQAWholeDeltaCIHaiku}{\mbox{[$-$6.1, $-$3.2]}}
\newcommand{\PlotQAHorizontalDeltaHaiku}{$-$19.2}
\newcommand{\PlotQAHorizontalDeltaCIHaiku}{\mbox{[$-$22.6, $-$15.9]}}
\newcommand{\PlotQAOtherDeltaHaiku}{$-$6.8}
\newcommand{\PlotQAOtherDeltaCIHaiku}{\mbox{[$-$12.6, $-$0.9]}}
\newcommand{\PlotQABestGptNano}{56.7}
\newcommand{\PlotQAHorizontalGptNano}{51.8}
\newcommand{\PlotQAOtherGptNano}{77.1}
\newcommand{\PlotQAWholeGptNano}{25.3}
\newcommand{\PlotQADeltaGptNano}{$-$30.3}
\newcommand{\PlotQADeltaCIGptNano}{\mbox{[$-$33.3, $-$27.3]}}
\newcommand{\PlotQAWholeDeltaGptNano}{$-$11.2}
\newcommand{\PlotQAWholeDeltaCIGptNano}{\mbox{[$-$12.5, $-$9.8]}}
\newcommand{\PlotQAHorizontalDeltaGptNano}{$-$34.5}
\newcommand{\PlotQAHorizontalDeltaCIGptNano}{\mbox{[$-$37.9, $-$31.2]}}
\newcommand{\PlotQAOtherDeltaGptNano}{$-$12.5}
\newcommand{\PlotQAOtherDeltaCIGptNano}{\mbox{[$-$18.6, $-$6.9]}}
\newcommand{\PlotQABestDePlot}{87.0}
\newcommand{\PlotQAHorizontalDePlot}{86.3}
\newcommand{\PlotQAOtherDePlot}{89.6}
\newcommand{\PlotQAWholeDePlot}{36.5}
\newcommand{\PlotQATopNames}{Gemini 3 Flash and Claude Sonnet 4.6}
\newcommand{\PlotQATopDeltas}{+2.2 \mbox{[+0.3, +4.1]} and +2.1 \mbox{[$-$0.06, +4.2]}}
\newcommand{\PlotQATopPairDelta}{+0.1}
\newcommand{\PlotQATopPairDeltaCI}{\mbox{[$-$1.6, +1.8]}}
\newcommand{\PlotQABelowCount}{four}
\newcommand{\PlotQABelowList}{Gemini 3.1 Flash-Lite, GPT-5.4 mini, Claude Haiku 4.5 and GPT-5.4 nano}
\newcommand{\PlotQABelowMin}{3.8}
\newcommand{\PlotQABelowMax}{30.3}
\newcommand{\PlotQAOtherBorderlineName}{Gemini 3.1 Flash-Lite}
\newcommand{\PlotQAOtherBorderlineEnd}{0.03}
\newcommand{\PlotQAHorizontalTopDeltas}{+1.5 \mbox{[$-$0.6, +3.7]} and +1.5 \mbox{[$-$0.9, +3.9]}}
\newcommand{\PlotQAOtherTopDeltas}{+5.0 \mbox{[+1.0, +9.6]} and +4.5 \mbox{[$-$0.2, +9.4]}}
\newcommand{\PlotQAHorizontalShareTop}{55\% and 58\%}
\newcommand{\PlotQAHorizontalItemShare}{81\%}
\newcommand{\PlotQAOtherAboveList}{Gemini 3 Flash}
\newcommand{\PlotQAOtherLevelCount}{three}
\newcommand{\PlotQAOtherLevelList}{Claude Sonnet 4.6, Gemini 3.1 Flash-Lite and GPT-5.4 mini}
\newcommand{\PlotQAOtherBelowCount}{two}
\newcommand{\PlotQAOtherBelowList}{Claude Haiku 4.5 and GPT-5.4 nano}
\newcommand{\PlotQAOrientationGapMin}{6.3}
\newcommand{\PlotQAOrientationGapMax}{25.3}
\newcommand{\PlotQAOrientationGapDePlot}{3.3}
\newcommand{\PlotQAOrientationGapCIDePlot}{\mbox{[$-$2.7, +8.9]}}
\newcommand{\PlotQAWholeMin}{25}
\newcommand{\PlotQAWholeMax}{38}
\newcommand{\PlotQAWholeMovedName}{Gemini 3.1 Flash-Lite}
\newcommand{\PlotQAWholeMovedDelta}{$-$0.6 \mbox{[$-$1.6, +0.5]}}
\newcommand{\PlotQATopMarginMin}{0.8}
\newcommand{\PlotQATopMarginMax}{3.3}
\newcommand{\PlotQAOtherLevelWeakName}{GPT-5.4 mini}
\newcommand{\PlotQAOtherLevelWeakDelta}{+0.1 \mbox{[$-$3.6, +4.2]}}
\newcommand{\PlotQARepeatsTopDeltas}{+3.3 \mbox{[+1.5, +5.1]} and +2.8 \mbox{[+0.7, +5.0]}}
\newcommand{\AliasPlotQARunCount}{two}
\newcommand{\AliasPlotQAMediumBest}{74.3}
\newcommand{\AliasPlotQAMediumDelta}{$-$12.7 \mbox{[$-$15.2, $-$10.2]}}
\newcommand{\AliasPlotQASmallN}{511}
\newcommand{\AliasPlotQASmallBest}{72.7}
\newcommand{\AliasPlotQASmallDelta}{$-$14.2 \mbox{[$-$17.1, $-$11.3]}}
\newcommand{\PlotQARepeatsMovedDelta}{$-$1.7 \mbox{[$-$3.9, +0.4]}}
\newcommand{\PlotQAAlwaysBelowCount}{three}
\newcommand{\PlotQAAlwaysBelowList}{GPT-5.4 mini, Claude Haiku 4.5 and GPT-5.4 nano}
\newcommand{\PlotQARepeatsGainDePlot}{0.4}
\newcommand{\PlotQARepeatsGainMax}{5.2}
\newcommand{\PlotQARepeatsGainMaxName}{Claude Haiku 4.5}
\newcommand{\PlotQASmallN}{426}
\newcommand{\PlotQALargeN}{103}
\newcommand{\PlotQALargeExponent}{6}
\newcommand{\PlotQASmallTopDeltas}{$-$0.1 \mbox{[$-$1.3, +1.2]} and $-$2.0 \mbox{[$-$3.6, $-$0.4]}}
\newcommand{\PlotQALargeTopDeltas}{+11.4 \mbox{[+3.4, +19.8]} and +18.9 \mbox{[+11.0, +27.1]}}
\newcommand{\PlotQALargeShareTop}{102\% and 178\%}
\newcommand{\PlotQALargeHorizontalN}{83}
\newcommand{\PlotQALargeOtherN}{20}
\newcommand{\PlotQAThresholdLowExponent}{3}
\newcommand{\PlotQAThresholdHighExponent}{9}
\newcommand{\PlotQAMedErrDePlot}{0.05}
\newcommand{\PlotQAMedErrVLMMin}{0.12}
\newcommand{\PlotQAMedErrVLMMax}{0.98}
\newcommand{\PlotQADePlotShortReplies}{71}
\newcommand{\PlotQADePlotShortScore}{78.7}
\newcommand{\PlotQADePlotRestScore}{88.3}

\newcommand{\PromptDetailedSonnet}{88.6}
\newcommand{\PromptDeltaSonnet}{$-$0.5}
\newcommand{\PromptDeltaCISonnet}{\mbox{[$-$1.2, +0.3]}}
\newcommand{\PromptDeltaHighSonnet}{0.3}
\newcommand{\PromptDetailedHaiku}{69.7}
\newcommand{\PromptDeltaHaiku}{$-$0.4}
\newcommand{\PromptDeltaCIHaiku}{\mbox{[$-$2.0, +1.1]}}
\newcommand{\PromptDeltaHighHaiku}{1.1}

\newcommand{\TolChartXOneGeminiFlash}{94.3}
\newcommand{\TolChartXOneGeminiFlashLite}{93.4}
\newcommand{\TolChartXOneGptMini}{90.8}
\newcommand{\TolChartXOneSonnet}{86.8}
\newcommand{\TolChartXOneGptNano}{85.2}
\newcommand{\TolChartXOneHaiku}{81.9}
\newcommand{\TolChartXOneMistralMedium}{79.3}
\newcommand{\TolChartXOneMistralSmall}{71.3}
\newcommand{\TolChartXOneMistralLarge}{67.9}
\newcommand{\TolChartXOneDePlot}{68.2}
\newcommand{\TolPlotQAOneGeminiFlash}{82.8}
\newcommand{\TolPlotQAOneSonnet}{74.8}
\newcommand{\TolPlotQAOneGeminiFlashLite}{74.3}
\newcommand{\TolPlotQAOneGptMini}{53.3}
\newcommand{\TolPlotQAOneHaiku}{40.4}
\newcommand{\TolPlotQAOneGptNano}{31.7}
\newcommand{\TolPlotQAOneDePlot}{82.8}
\newcommand{\TolChartXTwoGeminiFlash}{95.0}
\newcommand{\TolChartXTwoGeminiFlashLite}{94.1}
\newcommand{\TolChartXTwoGptMini}{91.9}
\newcommand{\TolChartXTwoSonnet}{89.1}
\newcommand{\TolChartXTwoGptNano}{86.3}
\newcommand{\TolChartXTwoHaiku}{84.2}
\newcommand{\TolChartXTwoMistralMedium}{80.9}
\newcommand{\TolChartXTwoMistralSmall}{74.7}
\newcommand{\TolChartXTwoMistralLarge}{71.4}
\newcommand{\TolChartXTwoDePlot}{71.3}
\newcommand{\TolPlotQATwoGeminiFlash}{86.0}
\newcommand{\TolPlotQATwoSonnet}{83.0}
\newcommand{\TolPlotQATwoGeminiFlashLite}{79.1}
\newcommand{\TolPlotQATwoGptMini}{64.9}
\newcommand{\TolPlotQATwoHaiku}{55.9}
\newcommand{\TolPlotQATwoGptNano}{42.9}
\newcommand{\TolPlotQATwoDePlot}{85.1}
\newcommand{\TolChartXFiveGeminiFlash}{96.0}
\newcommand{\TolChartXFiveGeminiFlashLite}{95.1}
\newcommand{\TolChartXFiveGptMini}{93.4}
\newcommand{\TolChartXFiveSonnet}{92.2}
\newcommand{\TolChartXFiveGptNano}{89.0}
\newcommand{\TolChartXFiveHaiku}{88.7}
\newcommand{\TolChartXFiveMistralMedium}{84.8}
\newcommand{\TolChartXFiveMistralSmall}{80.2}
\newcommand{\TolChartXFiveMistralLarge}{79.1}
\newcommand{\TolChartXFiveDePlot}{74.3}
\newcommand{\TolPlotQAFiveGeminiFlash}{89.2}
\newcommand{\TolPlotQAFiveSonnet}{89.0}
\newcommand{\TolPlotQAFiveGeminiFlashLite}{83.2}
\newcommand{\TolPlotQAFiveGptMini}{74.0}
\newcommand{\TolPlotQAFiveHaiku}{70.2}
\newcommand{\TolPlotQAFiveGptNano}{56.7}
\newcommand{\TolPlotQAFiveDePlot}{87.0}
\newcommand{\TolChartXTenGeminiFlash}{96.5}
\newcommand{\TolChartXTenGeminiFlashLite}{95.9}
\newcommand{\TolChartXTenGptMini}{94.2}
\newcommand{\TolChartXTenSonnet}{94.2}
\newcommand{\TolChartXTenGptNano}{90.8}
\newcommand{\TolChartXTenHaiku}{91.6}
\newcommand{\TolChartXTenMistralMedium}{87.5}
\newcommand{\TolChartXTenMistralSmall}{83.8}
\newcommand{\TolChartXTenMistralLarge}{84.8}
\newcommand{\TolChartXTenDePlot}{76.0}
\newcommand{\TolPlotQATenGeminiFlash}{90.1}
\newcommand{\TolPlotQATenSonnet}{92.1}
\newcommand{\TolPlotQATenGeminiFlashLite}{85.7}
\newcommand{\TolPlotQATenGptMini}{77.2}
\newcommand{\TolPlotQATenHaiku}{75.4}
\newcommand{\TolPlotQATenGptNano}{64.6}
\newcommand{\TolPlotQATenDePlot}{87.7}
\newcommand{\TolChartXAboveOne}{eight}
\newcommand{\TolChartXMarginGrows}{six}
\newcommand{\TolChartXMarginShrinksList}{Claude Haiku 4.5, Mistral Small 4 and Mistral Large 3}
\newcommand{\TolChartXOneLowestDelta}{$-$0.3 \mbox{[$-$3.6, +3.0]}}
\newcommand{\TolChartXTwoLowestDelta}{+0.1 \mbox{[$-$3.0, +3.2]}}
\newcommand{\TolChartXOneSecondLowestName}{Mistral Small 4}
\newcommand{\TolChartXOneSecondLowestDelta}{+3.1 \mbox{[+0.09, +6.2]}}
\newcommand{\TolPlotQATenLevelName}{Gemini 3.1 Flash-Lite}
\newcommand{\TolPlotQATenLevelDelta}{$-$2.0 \mbox{[$-$4.0, +0.04]}}
\newcommand{\TolPlotQATieName}{Gemini 3 Flash}
\newcommand{\TolPlotQAOneSecondName}{Claude Sonnet 4.6}
\newcommand{\TolPlotQAOneSecondDelta}{$-$8.1 \mbox{[$-$10.7, $-$5.2]}}
\newcommand{\TolPlotQATenTopDeltas}{+2.5 \mbox{[+0.6, +4.4]} and +4.4 \mbox{[+2.5, +6.5]}}
\newcommand{\TolPlotQADropDoubled}{five}

\newcommand{\PMCPairsHaiku}{259}
\newcommand{\PMCRecallHaiku}{12.9}
\newcommand{\PMCPairsSonnet}{255}
\newcommand{\PMCRecallSonnet}{13.6}
\newcommand{\PMCMedianCells}{71}
\newcommand{\PMCPairsOpus}{13}

\begin{document}

\maketitle

\begin{abstract}
Systematic reviews and meta-analyses often need numerical data reported only in figures, and extracting them with interactive digitisers usually requires a person to select and calibrate each figure. We present PlotPick, an open-source tool that uses vision-language models (VLMs) to extract tabular data from batches of scientific figures, and we benchmark the kind of model it calls: \ChartXModelCount{} VLMs from \ProviderCount{} providers on ChartX and \PlotQAModelCount{} of them on PlotQA, against DePlot, a dedicated chart-to-table model, with every system scored on the same items by numeric F1 (F1 over unlabelled numbers at \Tolerance\% relative tolerance). On \ChartXTypeCount{} ChartX chart types ($n{=}\ChartXNDePlot$) all \ChartXModelCount{} VLMs outperform DePlot in aggregate, at \ChartXFoneVLMMin--\ChartXFoneVLMMax\% against \ChartXFoneDePlot\%. The lead comes mainly from box plots, where DePlot scores \ChartXBoxDePlot\% against \ChartXBoxVLMMin--\ChartXBoxVLMMax\%; pooled over the other \ChartXNonBoxTypeCount{} types, \ChartXSweepCount{} VLMs keep a lead of \ChartXNonBoxLeadMin{} to \ChartXNonBoxLeadMax{} points and the \ChartXTrailingModels{} weakest do not. On a subset of the PlotQA test split ($n{=}\PlotQAN$; \PlotQAHorizontalN{} horizontal bar charts), scored by a lenient best-series variant of the metric, DePlot reaches \PlotQABestDePlot\%; the two strongest VLMs are level with it or slightly above it, and \PlotQABelowCount{} fall \PlotQABelowMin{} to \PlotQABelowMax{} points below. DePlot was trained on PlotQA's training split. Both benchmarks use synthetic charts, and the metric ignores which series a value belongs to. The application itself was not evaluated: its figure detection, structured output, prompt and default model were not tested, and the one benchmarked model it offers, Claude Haiku 4.5, is one of the \PlotQABelowCount{} below DePlot on PlotQA. Accuracy on biomedical figures has not been established, and every extracted value must be checked against its source figure. This version corrects version 1, which scored most PlotQA replies against category labels instead of plotted values; its claim that every VLM outperformed DePlot on both benchmarks is withdrawn. PlotPick is available at \url{https://plotpick.streamlit.app/}.
\end{abstract}

\section{Introduction}

Data locked in figures is a recurring obstacle in evidence synthesis~\cite{pick2019metadigitise}. In a two-rater study of figures from randomised trials, extraction with the open-source Plot Digitizer took 47\% less time than manual estimation and agreed with the trial authors' source data somewhat more often (75\% and 73\% for the two raters, against 66\% and 69\%)~\cite{jelicickadic2016}. Interactive digitisers such as WebPlotDigitizer~\cite{rohatgi_wpd} and the R packages metaDigitise~\cite{pick2019metadigitise} and juicr~\cite{lajeunesse2026juicr} export structured data, and each can save a record of an extraction so that it can be reproduced. WebPlotDigitizer has been evaluated for intercoder reliability and validity in an independent study of single-case-design graphs~\cite{drevon2017}; PlotPick has had no such evaluation. None of these three tools finds the figures in a PDF. A person supplies or selects each figure and calibrates its axes, with two exceptions that we did not test: juicr's automated mode for scatter and bar plots, and the AI Assist of WebPlotDigitizer, an experimental closed-source feature described as using multimodal models to detect a chart's axes and data~\cite{rohatgi_wpd}.

Recent work has shown that general-purpose multimodal models can extract plot data without task-specific training: PlotExtract~\cite{polak2025plotextract} reports over 90\% precision and around 90\% recall on the two-axis plots its workflow identifies as extractable, using engineered zero-shot prompts without fine-tuning. That vision-language models (VLMs) can read charts is therefore not our contribution. Ours is a comparison of \ChartXModelCount{} VLMs with a dedicated chart-to-table model under one metric on the same items, and an application that applies the approach to batches of PDFs.

Models specialised for charts are trained for this task. DePlot~\cite{liu2023deplot} translates a plot into a table and does nothing else; TinyChart~\cite{zhang2024tinychart} is a larger chart-understanding model with chart-to-table among its tasks. DePlot's plot-to-table training data is an equal mix of synthetic charts, ChartQA and PlotQA, and its authors describe line, dot, bar and pie charts~\cite{liu2023deplot}. The authors of ChartX already report GPT-4V above DePlot on their structural-extraction task~\cite{chartx2024}. We compare the VLMs with DePlot on two benchmarks: \ChartXTypeCount{} chart types of ChartX~\cite{chartx2024}, and a subset of the test split of PlotQA~\cite{methani2020plotqa}, one of DePlot's training sources. The two give different answers. On ChartX all \ChartXModelCount{} VLMs lead DePlot in aggregate, mainly because of box plots, a type not among those DePlot's authors describe. On PlotQA the two strongest VLMs are level with DePlot or slightly above it, and DePlot is ahead of the other \PlotQABelowCount{} (Section~\ref{sec:evaluation}). We ran no chart-specific model besides DePlot (TinyChart, for example, was not run), so our results speak to DePlot only.

PlotPick combines automatic figure detection from PDFs with VLM-based extraction, requiring no model training or manual annotation. The benchmarks evaluate VLMs of the kind the application calls, not the application itself: of the \ChartXModelCount, it offers only Claude Haiku 4.5 (Section~\ref{sec:design}).

\medskip
\noindent Hosted demo: \url{https://plotpick.streamlit.app/} \\
Source code: \url{https://github.com/tommycarstensen/plotpick}
\medskip

This version replaces version 1, several of whose claims were wrong. Section~\ref{sec:changes} lists what changed and why.

\section{Software design}
\label{sec:design}

PlotPick is a small Streamlit\footnote{\url{https://streamlit.io}} application whose figure-detection and model-configuration logic lives in separate modules. Uploaded PDFs are parsed with pypdfium2~\cite{pypdfium2}, a Python binding to PDFium; the characters it reports are grouped into text blocks with their bounding boxes. A block whose text matches a caption pattern (for example ``Figure 1'' or ``Table 2'') anchors a crop region. For a figure caption the region is expanded to enclose the raster images that belong to that caption or, where there are none, the vector graphics above it; for a table caption it runs down through the text blocks under the caption. The crop includes the caption, and a page on which no caption is found is offered whole. Each crop is then sent to a vision-language model, and the returned table is displayed alongside the source figure with a model-reported confidence score per figure and highlighting of the fields the model flagged as uncertain. The confidence score is self-reported, and we did not test whether it, or the flags, identify the figures or values that are wrong.

Four details matter for interpreting the results below. First, the application sends a long structured prompt that asks for JSON keyed to fields a meta-analyst would use (biomarker, group and timepoint for every chart type; median and quartiles for box plots; group size, mean, error and error type for bar charts; mean or median, error and error type for line charts). The set is not complete: box-plot and line-chart rows carry no group size, and the line-chart field does not record whether a value is a mean or a median. The prompt tells the model to interpolate values from the nearest axis ticks and to return a row for every box, bar or point even when a value is approximate. The benchmarks in Section~\ref{sec:evaluation}, by contrast, use a two-sentence prompt (``Extract the data from this chart as a tab-separated table. Return ONLY the table, no explanation.'')\ so that every VLM is compared on equal terms and the result is not specific to our prompt. Second, the benchmarks score numeric values only; the structured fields, the confidence score and the figure detection were not evaluated. Third, the application's default model, Claude Sonnet 5.5, and its optional Claude Opus 5.5 were not benchmarked; of the models the application offers, only Claude Haiku 4.5 was. Fourth, the application sends images up to 2,000 pixels wide, where the benchmarks sent at most \ImageMaxSide{} pixels on the longer side; we did not test whether the larger images help or hurt. The benchmark numbers therefore characterise VLMs of the kind the application calls, and not the application as shipped.

\needspace{6\baselineskip}
Key design choices:

\begin{itemize}
  \item \textbf{Model-agnostic in principle}: the extraction step depends only on a model accepting an image and returning text, and we benchmark \ChartXModelCount{} models from \ProviderCount{} providers on that basis. The released application ships an Anthropic backend only; another provider would need new code for its API, which we have not written.
  \item \textbf{Batch processing}: multiple figures can be processed in a single session, with results accumulated across uploads.
  \item \textbf{Export formats}: Markdown, Excel, CSV, LaTeX, JSON and R scripts. The tabular exports contain the extracted rows only; the model's per-row uncertainty flags, which are highlighted on screen, and the confidence score are included in the JSON export alone.
\end{itemize}

\section{Evaluation}
\label{sec:evaluation}

We evaluated \ChartXModelCount{} VLMs from \ProviderCount{} providers (Table~\ref{tab:models}) through their APIs on two public chart-to-table benchmarks and compared them with DePlot, a dedicated chart-to-table model with 282M parameters~\cite{liu2023deplot}. Every VLM run in the tables and figures used the same two-sentence prompt with no model-specific tuning, at its provider's defaults for sampling and for reasoning, which differ between models and were not controlled, and without a seed. The Claude and OpenAI calls capped each reply at 2,000 output tokens (for the OpenAI models, any reasoning tokens included), and whether a reply stopped at that cap was not recorded (the released runner allows more for Claude; see its README); the Gemini and Mistral calls used their providers' default limits. Each image sent to a VLM was downscaled to at most \ImageMaxSide{} pixels on its longer side.

DePlot was run from its public checkpoint\footnote{\url{https://huggingface.co/google/deplot}} on the original image, which its processor rescales to its own patch budget, with the instruction of its model card (``Generate underlying data table of the figure below:''), greedy decoding and at most \DePlotMaxNewTokens{} new tokens. A reply that stops at that limit has \DePlotMaxNewTokens{} tokens, and an ordinary token decodes to at least one character, so only a reply of \DePlotMaxNewTokens{} characters or longer could have reached it: \ChartXDePlotLongReplies{} of its \ChartXNDePlot{} ChartX replies are that long, and none of its PlotQA replies. On the development split, \DevOtherDePlotLongReplies{} of its \DevOtherN{} replies on the \ChartXTypesOther{} unreported types are that long; leaving them out moves the margins quoted for those types in Section~\ref{sec:chartx} by less than a point. DePlot was run on PlotQA in April 2026 and on the ChartX validation split in October 2026.

Each system was run once per benchmark and prompt. A call that returned an empty reply was repeated, and a reply that stayed empty was removed from the results instead of being scored as zero; Section~\ref{sec:chartx} gives the ChartX scores with those replies counted. No analysis plan was registered. The chart types, the tolerance and the scoring were chosen by the author after results on the ChartX development split (Section~\ref{sec:chartx}) had been seen, and the scoring was corrected during the preparation of this version (Section~\ref{sec:changes}).

\begin{table}[ht]
  \centering
  \caption{VLMs evaluated, with the identifier sent to each provider's API. The \NonMistralModelCount{} non-Mistral models were run in April and May 2026 and the \MistralModelCount{} Mistral models in October 2026. The two Gemini identifiers are preview versions; Google's deprecation schedule gives 25 May 2026 as the shutdown date of \texttt{gemini-3.1-flash-lite-preview}, so that run cannot be repeated under the same identifier. The two OpenAI identifiers carry no date; OpenAI's documentation lists one snapshot for each, dated 17 March 2026.}
  \label{tab:models}
  \small
  \begin{tabular}{llll}
    \toprule
    Provider  & Model & API identifier & Benchmarks \\
    \midrule
    Google    & Gemini 3 Flash (preview)        & \texttt{gemini-3-flash-preview}        & ChartX, PlotQA \\
    Google    & Gemini 3.1 Flash-Lite (preview) & \texttt{gemini-3.1-flash-lite-preview} & ChartX, PlotQA \\
    OpenAI    & GPT-5.4 mini                    & \texttt{gpt-5.4-mini}                  & ChartX, PlotQA \\
    OpenAI    & GPT-5.4 nano                    & \texttt{gpt-5.4-nano}                  & ChartX, PlotQA \\
    Anthropic & Claude Sonnet 4.6               & \texttt{claude-sonnet-4-6}             & ChartX, PlotQA \\
    Anthropic & Claude Haiku 4.5                & \texttt{claude-haiku-4-5-20251001}     & ChartX, PlotQA \\
    Mistral   & Mistral Medium 3.5              & \texttt{mistral-medium-2604}           & ChartX \\
    Mistral   & Mistral Small 4                 & \texttt{mistral-small-2603}            & ChartX \\
    Mistral   & Mistral Large 3                 & \texttt{mistral-large-2512}            & ChartX \\
    \bottomrule
  \end{tabular}
\end{table}

The PlotQA results cover \PlotQAModelCount{} of the \ChartXModelCount{} models. The \MistralModelCount{} Mistral models were added for this version and run on ChartX only. Earlier Mistral runs used moving \texttt{-latest} aliases, which cannot be tied to a model version; they are kept in the released results and are not tabulated. Sections~\ref{sec:chartx} and~\ref{sec:plotqa} say how those on the ChartX validation split and on PlotQA score.

\subsection{Metrics}
\label{sec:metrics}

We report two metrics. \textbf{Recall} is the fraction of ground-truth numeric values for which the extraction contains a value within \Tolerance\% relative tolerance. It carries no precision term, so a system that emits many values is rewarded for chance matches. \textbf{Numeric F1} is the harmonic mean of precision and recall when extracted and ground-truth values are matched one to one under the same tolerance, and is our headline metric. Both work on sets of numbers, so a value that occurs twice in a table counts once, and both count digits in headers and row labels like any other number (the best-series score of Section~\ref{sec:plotqa} grades each column without its header cell and each row without its first cell). Sections~\ref{sec:chartx} and~\ref{sec:plotqa} show what these two choices cost.

Version 1 of this paper called numeric F1 ``RMSF1''. Numeric F1 is not the Relative Mapping Similarity F1 (RMS F1) of the DePlot paper~\cite{liu2023deplot}, which matches (row header, column header, value) triples with partial credit. Numeric F1 is closer to the numbers-only metric that the DePlot paper calls relative number set similarity (RNSS) and attributes to ChartQA~\cite{masry2022chartqa}. RNSS grades each matched pair by its relative error, whereas numeric F1 applies a hard tolerance. Our scores are therefore not comparable with published RMS F1 or RNSS figures. We report no RMS F1 because only the numbers extracted from the VLM replies were stored for the ChartX validation split, and the PlotQA annotations we used hold one series per chart.

Both metrics compare unlabelled sets of numbers and therefore do not check that a value was assigned to the correct series, group or timepoint. They measure whether the right numbers were read, not whether the right table was produced. The tolerance is also wide for many purposes: a true mean of \ToleranceExample{} is accepted anywhere from \ToleranceExampleLow{} to \ToleranceExampleHigh. Section~\ref{sec:tolerance} repeats the main comparisons at tighter and looser tolerances.

DePlot begins every reply with a title line. We remove it before scoring, because the VLMs were asked for the table alone and a number in a title would otherwise count as a wrongly extracted value; \ChartXDePlotTitlesWithNumber{} of DePlot's \ChartXNDePlot{} ChartX titles contain one. Counting the title line would lower DePlot's ChartX score from \ChartXFoneDePlot\% to \ChartXFoneDePlotTitled\%. A title line in a VLM reply on that split could not have been removed, because the replies themselves were not stored; any such line counts against the VLM.

Numbers are found in each reply with a single regular expression, applied alike to every system. It reads a hyphen directly before a digit as a minus sign and does not expand unit suffixes, so a range, a date or a value written as ``3.2M'' is misread.

Intervals are 95\% percentile bootstrap intervals from \BootstrapDraws{} resamples of items. Because all systems were scored on the same items, a difference between two systems is bootstrapped from the per-item differences (a paired interval). We write it as the difference in percentage points followed by its interval in square brackets, and call the difference separable when the interval excludes zero. Two comparisons are between different sets of charts (charts with and without data labels, and the two orientations of the PlotQA charts); for those the two sets are resampled independently (an unpaired interval). Intervals are not adjusted for the number of comparisons, and they reflect the sampling of items, not the variation between repeated runs of a model. Differences are computed from unrounded scores on the items two systems share, so they can differ in the last digit from differences of the rounded values in the tables. An interval end within \BorderlinePoints{} points of zero is given to two decimals; whether such an interval excludes zero could change with the bootstrap seed.

\subsection{ChartX benchmark}
\label{sec:chartx}

ChartX~\cite{chartx2024} includes \ChartXTypesTotal{} chart types. We report \ChartXTypeCount{} that are common in scientific publications (plain bar chart, bar chart with data labels, plain line chart, line chart with data labels, box plot, histogram). The reported results are on the validation file of the released dataset (\ChartXValItems{} items), from which we took the first \ChartXPerType{} items of each type in the order of the file, \ChartXNDePlot{} per system; the \ChartXTestItems-item test file served as our development split. ChartX's authors report on the test file instead~\cite{chartx2024}. We did not train or fine-tune any system on ChartX, and whether the VLMs saw it in their own training is unknown (Section~\ref{sec:limitations}); our numbers are not comparable with those of the ChartX authors. The validation items were not audited for ground-truth errors.

Claude Haiku 4.5 returned an empty reply for one plain bar chart, which was removed from its results, so it is scored on \ChartXNHaiku{} items; comparisons involving it pair the \ChartXNHaiku{} shared items, on which DePlot scores \ChartXFoneDePlotHaikuItems\%. Scored as zero, that reply would give Claude Haiku 4.5 \ChartXFoneHaikuZero\% on all \ChartXNDePlot{} items and a margin over DePlot of \ChartXDeltaHaikuZero~points.

The \ChartXTypeCount{} types were chosen after development runs that covered all \ChartXTypesTotal. For the other \ChartXTypesOther{} types the only VLM runs are those of the two Claude models on the development split. On the \DevOtherN{} items for which DePlot was run there, DePlot scores \DevOtherDePlot\%, and Claude Haiku 4.5 and Claude Sonnet 4.6 lead it by \DevOtherDeltaHaiku~\DevOtherDeltaCIHaiku{} and \DevOtherDeltaSonnet~\DevOtherDeltaCISonnet{} points (\DevOtherNHaiku{} and \DevOtherNSonnet{} shared items). On the \ChartXTypeCount{} reported types of the same split the two lead it by \DevSixDeltaHaiku~\DevSixDeltaCIHaiku{} and \DevSixDeltaSonnet~\DevSixDeltaCISonnet{} points (\DevSixN{} items). The choice of types therefore did not enlarge the lead of these two models over DePlot, but it did raise their scores: on the development split, with the longer prompt, Claude Haiku 4.5 averages \DevSixFoneHaiku\% numeric F1 over the six reported types and \DevOtherFoneHaiku\% over the other \ChartXTypesOther, Claude Sonnet 4.6 \DevSixFoneSonnet\% and \DevOtherFoneSonnet\%, and DePlot \DevSixFoneDePlot\% and \DevOtherFoneDePlot\% (means of the per-type scores). The other VLMs were not run on the other types. On the \ChartXTypeCount{} reported types, the \ChartXDevModels{} VLMs that were run on both splits score \ChartXDevGapMin{} to \ChartXDevGapMax{} points higher on the development split than on the validation split (mean of the per-type scores). The Claude development runs used a longer prompt than the validation runs.

During development we identified errors in the ground truth of \ChartXCorrections{} ChartX figures: CSV columns not depicted in the chart, redrawing scripts with typos or undefined variables, and stacked-bar annotations reporting totals where the segments are the data. We proposed all \ChartXCorrections{} upstream as pull requests on the dataset repository,\footnote{\url{https://huggingface.co/datasets/InternScience/ChartX/discussions}} which were unmerged at the time of writing. Locally we removed the undepicted columns from the affected tables and regenerated some of the images; the released benchmark repository lists, with links, what was changed for each figure. The errors were found by reviewing figures on which several models disagreed with the ground truth, so the search was not blind to model output; each was then verified by eye against the rendered chart, and no change was made merely because a model's answer looked preferable. All \ChartXCorrections{} figures are in the development split and none is in the validation split, so Tables~\ref{tab:chartx} and~\ref{tab:tolerance} and both figures are computed on unmodified upstream ground truth.

\begin{table}[ht]
  \centering
  \caption{Number extraction on the ChartX validation split, \ChartXTypeCount{} chart types, \ChartXPerType{} items per type. Numeric F1 and recall are at \Tolerance\% relative tolerance, and every system was scored by the same code from deduplicated value sets. The first interval is that of the system's own score. The margin is the system's numeric F1 minus DePlot's, paired over shared items, in percentage points, with its paired interval; Claude Haiku 4.5 is paired on \ChartXNHaiku{} items, on which DePlot scores \ChartXFoneDePlotHaikuItems\%. Rows are ordered by numeric F1.}
  \label{tab:chartx}
  \small
  \setlength{\tabcolsep}{5pt}
  \begin{tabular}{lrrlrrl}
    \toprule
    Model & $n$ & Numeric F1 & 95\% CI & Recall & Margin over DePlot & Paired 95\% CI \\
    \midrule
    Gemini 3 Flash        & \ChartXNGeminiFlash     & \ChartXFoneGeminiFlash\%     & \ChartXFoneCIGeminiFlash     & \ChartXRecGeminiFlash\%     & \ChartXDeltaGeminiFlash     & \ChartXDeltaCIGeminiFlash \\
    Gemini 3.1 Flash-Lite & \ChartXNGeminiFlashLite & \ChartXFoneGeminiFlashLite\% & \ChartXFoneCIGeminiFlashLite & \ChartXRecGeminiFlashLite\% & \ChartXDeltaGeminiFlashLite & \ChartXDeltaCIGeminiFlashLite \\
    GPT-5.4 mini          & \ChartXNGptMini         & \ChartXFoneGptMini\%         & \ChartXFoneCIGptMini         & \ChartXRecGptMini\%         & \ChartXDeltaGptMini         & \ChartXDeltaCIGptMini \\
    Claude Sonnet 4.6     & \ChartXNSonnet          & \ChartXFoneSonnet\%          & \ChartXFoneCISonnet          & \ChartXRecSonnet\%          & \ChartXDeltaSonnet          & \ChartXDeltaCISonnet \\
    GPT-5.4 nano          & \ChartXNGptNano         & \ChartXFoneGptNano\%         & \ChartXFoneCIGptNano         & \ChartXRecGptNano\%         & \ChartXDeltaGptNano         & \ChartXDeltaCIGptNano \\
    Claude Haiku 4.5      & \ChartXNHaiku           & \ChartXFoneHaiku\%           & \ChartXFoneCIHaiku           & \ChartXRecHaiku\%           & \ChartXDeltaHaiku           & \ChartXDeltaCIHaiku \\
    Mistral Medium 3.5    & \ChartXNMistralMedium   & \ChartXFoneMistralMedium\%   & \ChartXFoneCIMistralMedium   & \ChartXRecMistralMedium\%   & \ChartXDeltaMistralMedium   & \ChartXDeltaCIMistralMedium \\
    Mistral Small 4       & \ChartXNMistralSmall    & \ChartXFoneMistralSmall\%    & \ChartXFoneCIMistralSmall    & \ChartXRecMistralSmall\%    & \ChartXDeltaMistralSmall    & \ChartXDeltaCIMistralSmall \\
    Mistral Large 3       & \ChartXNMistralLarge    & \ChartXFoneMistralLarge\%    & \ChartXFoneCIMistralLarge    & \ChartXRecMistralLarge\%    & \ChartXDeltaMistralLarge    & \ChartXDeltaCIMistralLarge \\
    \midrule
    DePlot                & \ChartXNDePlot          & \ChartXFoneDePlot\%          & \ChartXFoneCIDePlot          & \ChartXRecDePlot\%          &                             & \\
    \bottomrule
  \end{tabular}
\end{table}

\begin{figure}[ht]
  \centering
  \includegraphics[width=\textwidth]{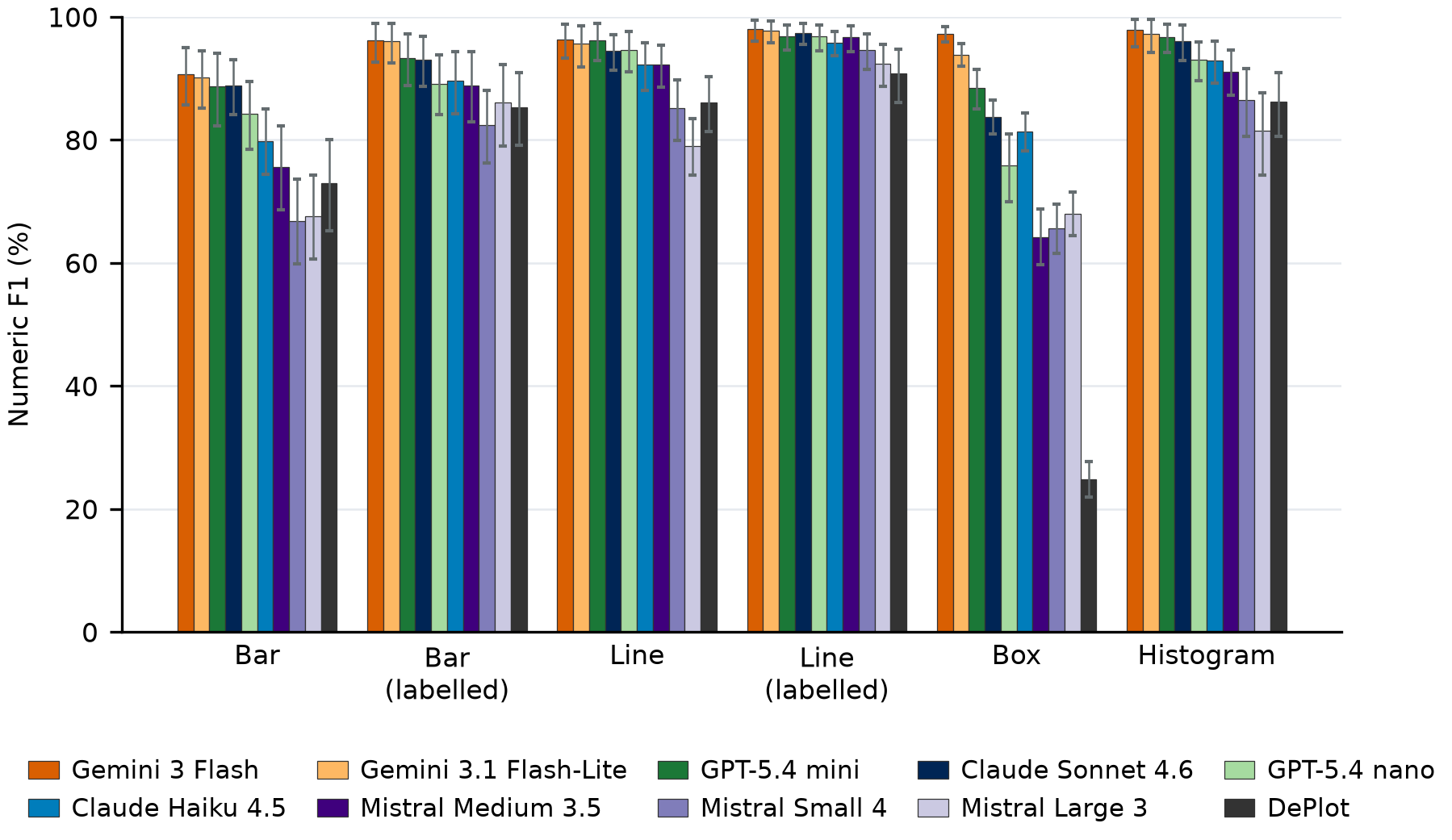}
  \caption{Numeric F1 (\Tolerance\% relative tolerance) by system and chart type on the ChartX validation split, \ChartXPerType{} items per type (one plain bar chart fewer for Claude Haiku 4.5). ``(labelled)'' marks charts that print their values as data labels. Bars follow the order of Table~\ref{tab:chartx}, with DePlot last. Error bars are 95\% bootstrap intervals for each system on its own; the comparisons in the text are paired, so two overlapping error bars do not mean that the paired interval of the difference includes zero.}
  \label{fig:results}
\end{figure}

All \ChartXModelCount{} VLMs outperform DePlot in aggregate (Table~\ref{tab:chartx}, Figure~\ref{fig:results}). The lowest-scoring VLM, \ChartXLowestVLMName, exceeds DePlot by \ChartXGapLowestVLM~points of numeric F1 (paired 95\% interval \ChartXGapLowestVLMCI), and every VLM also leads DePlot separably under recall. Not every adjacent pair of rows in Table~\ref{tab:chartx} is separable. The paired interval includes zero for \ChartXAdjacentNotSeparable. For \ChartXAdjacentBorderline{} the interval excludes zero by less than \BorderlinePoints~points, so the call could change with the bootstrap seed.

\AliasRunCountCap{} further runs are not tabulated because they were made under moving aliases. \AliasLeadCountCap{} of them lead DePlot separably, whether or not the empty replies of one of them, \texttt{mistral-\allowbreak small-\allowbreak latest}, are scored as zero. The other, \texttt{ministral-\allowbreak 3b-\allowbreak latest}, a smaller Mistral model, reached \MinistralFone\% numeric F1 on the \MinistralN{} items for which it returned a reply, where DePlot scores \MinistralDePlotFone\%. Its margin over DePlot is not separable on numeric F1 (\MinistralDelta~\MinistralDeltaCI) or on recall (\MinistralRecDelta~\MinistralRecDeltaCI), and the model is below DePlot on \MinistralTrailingTypes{} of the \ChartXTypeCount{} chart types. With its \MinistralMissing{} empty replies scored as zero it reaches \MinistralFoneZero\% on all \ChartXNDePlot{} items, \MinistralDeltaZero~\MinistralDeltaZeroCI{} against DePlot. ``All VLMs lead DePlot'' therefore describes the \ChartXModelCount{} models of Table~\ref{tab:chartx}, not every VLM.

\begin{figure}[ht]
  \centering
  \includegraphics[width=0.9\textwidth]{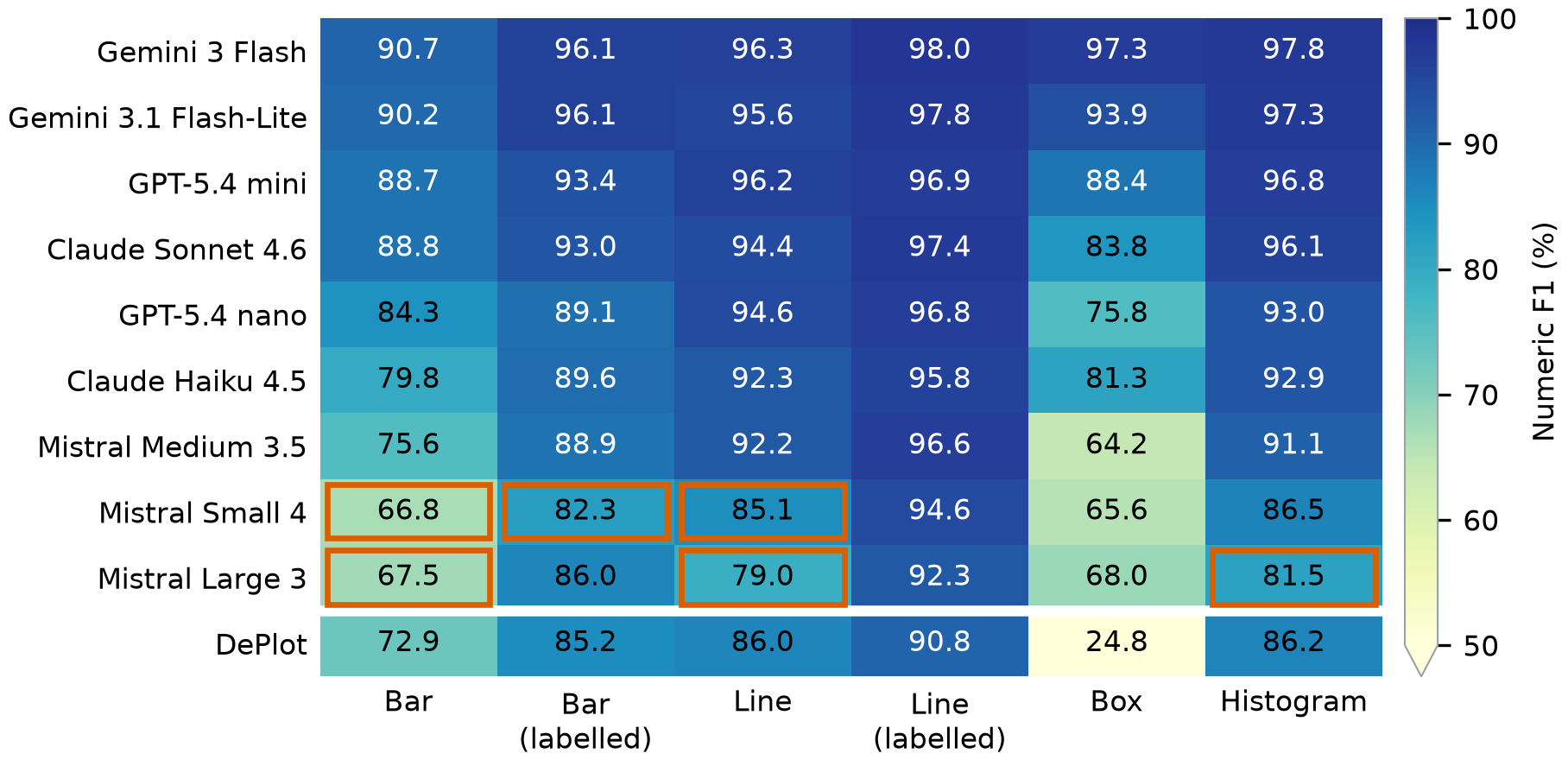}
  \caption{Numeric F1 (\%, at \Tolerance\% relative tolerance) by system and chart type, ChartX validation split, \ChartXPerType{} items per cell (one fewer for Claude Haiku 4.5 on plain bar charts). ``(labelled)'' marks charts that print their values as data labels. Rows follow the order of Table~\ref{tab:chartx}, with DePlot at the bottom. Orange outlines mark the \ChartXTrailingCells{} cells in which a VLM scores below DePlot: \ChartXTrailingList. Only \ChartXTrailingSeparable{} is separable from DePlot under a paired bootstrap. The colour scale starts at \HeatmapFloor\%, so DePlot's box-plot cell takes the lowest colour.}
  \label{fig:heatmap}
\end{figure}

The advantage is not uniform across chart types. By point estimate, \ChartXSweepCount{} of the \ChartXModelCount{} VLMs lead DePlot on every one of the \ChartXTypeCount{} types; the other \ChartXTrailingModels{} trail it in \ChartXTrailingCells{} of the \ChartXTotalCells{} model-by-type cells (Figure~\ref{fig:heatmap}): \ChartXTrailingList. At \ChartXPerType{} items per cell these counts are descriptive. Of the \ChartXTotalCells{} cells, \ChartXCellsLeadSeparable{} are separable in the VLM's favour, \ChartXCellsTrailSeparable{} in DePlot's (\ChartXTrailingSeparable) and \ChartXCellsSpanZero{} are not separable; those \ChartXCellsSpanZero{} include cells where the VLM leads by point estimate as well as cells where it trails. With \ChartXTotalCells{} unadjusted comparisons, one or two cells would be expected to be separable in each direction by chance, so the single cell in DePlot's favour is weak evidence, and so is any favourable cell whose interval ends close to zero. The number of cells below DePlot by point estimate also depends on the metric: under recall it is \ChartXTrailingCellsRecall.

The size of the margin depends on chart type. Averaged over the \ChartXModelCount{} VLMs it is \ChartXMarginBox~points on box plots, where DePlot scores \ChartXBoxDePlot\% against \ChartXBoxVLMMin--\ChartXBoxVLMMax\%, and between \ChartXMarginMeanMin{} and \ChartXMarginMeanMax~points on each of the other \ChartXNonBoxTypeCount{} types. DePlot's replies explain its box-plot score: it returns one value per box, on average \ChartXBoxDePlotNumbers{} numbers for a chart whose ground truth holds \ChartXBoxTruthNumbers{} distinct numbers (each box's minimum, quartiles, median and maximum, its outliers and the digits in its labels), so it would score at most \ChartXBoxDePlotCeiling\% even if every number it returned were right. On the other \ChartXNonBoxTypeCount{} types the lowest-scoring VLM of each type is between \ChartXMarginLowestMin~points behind DePlot and \ChartXMarginLowestMax~points ahead. The \ChartXTopCount{} strongest VLMs lead DePlot separably in every cell of those \ChartXNonBoxTypeCount{} types, by \ChartXTopCellMin{} to \ChartXTopCellMax~points. Pooled over the \ChartXNonBoxTypeCount{} types, \ChartXSweepCount{} VLMs keep a separable lead over DePlot, of \ChartXNonBoxLeadMin{} to \ChartXNonBoxLeadMax~points. Of the other \ChartXTrailingModels, one is not separable from DePlot (\ChartXNonBoxLevel) and one is below it (\ChartXNonBoxBehind); that interval would include zero if DePlot's title line were scored (\ChartXNonBoxBehindTitled). In this equal mix of \ChartXTypeCount{} types, box plots contribute \ChartXBoxShareMin--\ChartXBoxShareMax\% of the aggregate margin of the \ChartXSweepCount{} stronger models and more than all of it for the \ChartXTrailingModels{} weakest.

The metric also credits digits in labels. Numbers that occur only in a header or a row label (a year, the 0 and $-$10 that our regular expression reads from a histogram bin ``0-10'', the 1 and 3 of ``Q1'' and ``Q3'') make up \ChartXLabelShareMin--\ChartXLabelShareMax\% of the ground-truth numbers, depending on the chart type. Leaving them out of both the ground truth and the extraction lowers the margins of the VLMs by up to \ChartXLabelFallMax~points. The other VLMs keep a separable lead, but \ChartXLowestVLMName{} would lead DePlot by \ChartXLabelLowestDelta~\ChartXLabelLowestDeltaCI, which is not separable. The lead of the weakest VLM in Table~\ref{tab:chartx} therefore depends on counting label digits. The effect of counting each value once could be examined for DePlot only, the one system whose replies on this split are stored: \ChartXTruthRepeats{} of the \ChartXNDePlot{} ground-truth tables repeat a value, and counting repeats separately gives DePlot \ChartXFoneDePlotRepeats\% instead of \ChartXFoneDePlot\%.

\subsection{PlotQA benchmark}
\label{sec:plotqa}

PlotQA~\cite{methani2020plotqa} is one of the three sources of DePlot's training data. According to the DePlot paper, DePlot was trained on charts from the training splits of its sources only~\cite{liu2023deplot}, so the test-split charts used here should be unseen by it, though they are drawn by the same generator. We started from the first \PlotQASubsetN{} entries of the test split in the order of a public conversion of the dataset,\footnote{\url{https://huggingface.co/datasets/achang/plot_qa}} each reduced to the first series of its chart. These entries are not representative of PlotQA: \PlotQASubsetHorizontal{} are horizontal bar charts and \PlotQASubsetOther{} are line or dot-line plots, with no vertical bar chart, whereas horizontal bar charts are about a third of PlotQA's test split (11,292 of 33,657 images)~\cite{methani2020plotqa}. We ran DePlot ourselves and scored it with the same code as the VLMs.

Two things about these items were wrong in version 1 and are stated here plainly. The first is the ground truth. For a horizontal bar chart the annotation stores the category labels in one field and the bar lengths in another, and our runner read the same field for every chart. For \PlotQAHorizontalN{} of the \PlotQAN{} items that field held the category labels, which are years, so version 1 scored replies against a list of years. Every PlotQA score in this version is computed against the plotted values. The second is the selection. Our runner skipped the \PlotQASkipped{} entries for which the mistaken field was not numeric, so the \PlotQAN{} scored items are the \PlotQAHorizontalN{} horizontal bar charts whose categories are years and all \PlotQAOtherN{} other plots.

Because the annotation holds one series while every system returns the whole table, we score each column of a reply (header row excluded) and each row (first cell excluded) against that series and keep the best, and call the result \textbf{best-series} numeric F1. It is optimistic: the annotation chooses which series of the reply is graded, so a reply that reads every value correctly but assigns it to the wrong series still scores 100\%. Scoring the whole reply against the one series instead charges every system for correctly extracting the series the annotation omits; that \textbf{whole-table} score is \PlotQAWholeMin--\PlotQAWholeMax\%.

DePlot reaches \PlotQABestDePlot\% (Table~\ref{tab:plotqa}). The two strongest VLMs, \PlotQATopNames, score \PlotQATopDeltas{} points against it. The two margins are the same size and the two models are not separable from each other (\PlotQATopPairDelta~\PlotQATopPairDeltaCI); one interval just excludes zero and the other just includes it, and which side each falls on changes with the checks below. We therefore call both level with DePlot or slightly above it. The other \PlotQABelowCount{} VLMs are below DePlot by \PlotQABelowMin{} to \PlotQABelowMax~points (\PlotQABelowList). The two untabulated runs made under moving aliases are below DePlot as well: \texttt{mistral-\allowbreak medium-\allowbreak latest} scores \AliasPlotQAMediumBest\% (\AliasPlotQAMediumDelta) and \texttt{mistral-\allowbreak small-\allowbreak latest} \AliasPlotQASmallBest\% on the \AliasPlotQASmallN{} items for which its reply is stored (\AliasPlotQASmallDelta).

\begin{table}[ht]
  \centering
  \caption{Number extraction on the PlotQA subset ($n{=}\PlotQAN$, of which \PlotQAHorizontalN{} are horizontal bar charts): numeric F1 (\Tolerance\% relative tolerance) against the plotted values of the one annotated series, sorted by best-series score with DePlot, which we ran ourselves, in its rank position. Best-series is optimistic: the annotation chooses which column or row of the reply is graded. Whole-table scores the entire reply against the one series. Each margin is the system's score minus DePlot's, as a paired difference in percentage points. These scores are not comparable with those of Table~\ref{tab:chartx}. The \MistralModelCount{} Mistral models of Table~\ref{tab:models} were not run on PlotQA.}
  \label{tab:plotqa}
  \small
  \setlength{\tabcolsep}{5pt}
  \begin{tabular}{lrrlrrl}
    \toprule
          & \multicolumn{3}{c}{Best-series} & \multicolumn{3}{c}{Whole-table} \\
    \cmidrule(lr){2-4}\cmidrule(lr){5-7}
    Model & Score & Margin & Paired 95\% CI & Score & Margin & Paired 95\% CI \\
    \midrule
    Gemini 3 Flash        & \PlotQABestGeminiFlash\%     & \PlotQADeltaGeminiFlash     & \PlotQADeltaCIGeminiFlash     & \PlotQAWholeGeminiFlash\%     & \PlotQAWholeDeltaGeminiFlash     & \PlotQAWholeDeltaCIGeminiFlash \\
    Claude Sonnet 4.6     & \PlotQABestSonnet\%          & \PlotQADeltaSonnet          & \PlotQADeltaCISonnet          & \PlotQAWholeSonnet\%          & \PlotQAWholeDeltaSonnet          & \PlotQAWholeDeltaCISonnet \\
    DePlot                & \PlotQABestDePlot\%          &                             &                               & \PlotQAWholeDePlot\%          &                                  & \\
    Gemini 3.1 Flash-Lite & \PlotQABestGeminiFlashLite\% & \PlotQADeltaGeminiFlashLite & \PlotQADeltaCIGeminiFlashLite & \PlotQAWholeGeminiFlashLite\% & \PlotQAWholeDeltaGeminiFlashLite & \PlotQAWholeDeltaCIGeminiFlashLite \\
    GPT-5.4 mini          & \PlotQABestGptMini\%         & \PlotQADeltaGptMini         & \PlotQADeltaCIGptMini         & \PlotQAWholeGptMini\%         & \PlotQAWholeDeltaGptMini         & \PlotQAWholeDeltaCIGptMini \\
    Claude Haiku 4.5      & \PlotQABestHaiku\%           & \PlotQADeltaHaiku           & \PlotQADeltaCIHaiku           & \PlotQAWholeHaiku\%           & \PlotQAWholeDeltaHaiku           & \PlotQAWholeDeltaCIHaiku \\
    GPT-5.4 nano          & \PlotQABestGptNano\%         & \PlotQADeltaGptNano         & \PlotQADeltaCIGptNano         & \PlotQAWholeGptNano\%         & \PlotQAWholeDeltaGptNano         & \PlotQAWholeDeltaCIGptNano \\
    \bottomrule
  \end{tabular}
\end{table}

\begin{table}[ht]
  \centering
  \caption{The PlotQA subset by chart orientation: best-series numeric F1 (\Tolerance\% relative tolerance) and its paired margin over DePlot (the system's score minus DePlot's, in percentage points) on the horizontal bar charts and on the other plots, which are line or dot-line plots. Rows follow Table~\ref{tab:plotqa}.}
  \label{tab:orientation}
  \small
  \setlength{\tabcolsep}{5pt}
  \begin{tabular}{lrrlrrl}
    \toprule
          & \multicolumn{3}{c}{Horizontal bar charts ($n{=}\PlotQAHorizontalN$)} & \multicolumn{3}{c}{Other plots ($n{=}\PlotQAOtherN$)} \\
    \cmidrule(lr){2-4}\cmidrule(lr){5-7}
    Model & Score & Margin & Paired 95\% CI & Score & Margin & Paired 95\% CI \\
    \midrule
    Gemini 3 Flash        & \PlotQAHorizontalGeminiFlash\%     & \PlotQAHorizontalDeltaGeminiFlash     & \PlotQAHorizontalDeltaCIGeminiFlash     & \PlotQAOtherGeminiFlash\%     & \PlotQAOtherDeltaGeminiFlash     & \PlotQAOtherDeltaCIGeminiFlash \\
    Claude Sonnet 4.6     & \PlotQAHorizontalSonnet\%          & \PlotQAHorizontalDeltaSonnet          & \PlotQAHorizontalDeltaCISonnet          & \PlotQAOtherSonnet\%          & \PlotQAOtherDeltaSonnet          & \PlotQAOtherDeltaCISonnet \\
    DePlot                & \PlotQAHorizontalDePlot\%          &                                       &                                         & \PlotQAOtherDePlot\%          &                                  & \\
    Gemini 3.1 Flash-Lite & \PlotQAHorizontalGeminiFlashLite\% & \PlotQAHorizontalDeltaGeminiFlashLite & \PlotQAHorizontalDeltaCIGeminiFlashLite & \PlotQAOtherGeminiFlashLite\% & \PlotQAOtherDeltaGeminiFlashLite & \PlotQAOtherDeltaCIGeminiFlashLite \\
    GPT-5.4 mini          & \PlotQAHorizontalGptMini\%         & \PlotQAHorizontalDeltaGptMini         & \PlotQAHorizontalDeltaCIGptMini         & \PlotQAOtherGptMini\%         & \PlotQAOtherDeltaGptMini         & \PlotQAOtherDeltaCIGptMini \\
    Claude Haiku 4.5      & \PlotQAHorizontalHaiku\%           & \PlotQAHorizontalDeltaHaiku           & \PlotQAHorizontalDeltaCIHaiku           & \PlotQAOtherHaiku\%           & \PlotQAOtherDeltaHaiku           & \PlotQAOtherDeltaCIHaiku \\
    GPT-5.4 nano          & \PlotQAHorizontalGptNano\%         & \PlotQAHorizontalDeltaGptNano         & \PlotQAHorizontalDeltaCIGptNano         & \PlotQAOtherGptNano\%         & \PlotQAOtherDeltaGptNano         & \PlotQAOtherDeltaCIGptNano \\
    \bottomrule
  \end{tabular}
\end{table}

Three checks qualify this picture. The first is chart orientation (Table~\ref{tab:orientation}). On the \PlotQAHorizontalN{} horizontal bar charts neither of the two strongest VLMs is separable from DePlot (\PlotQAHorizontalTopDeltas) and the other \PlotQABelowCount{} are below it. On the \PlotQAOtherN{} other plots \PlotQAOtherAboveList{} is separably above DePlot, \PlotQAOtherLevelCount{} VLMs are not separable from it (\PlotQAOtherLevelList; the interval of \PlotQAOtherBorderlineName{} is only \PlotQAOtherBorderlineEnd{} points from zero) and \PlotQAOtherBelowCount{} are below it (\PlotQAOtherBelowList). The pooled result is therefore weighted towards the horizontal bar charts, which are \PlotQAHorizontalN{} of the \PlotQAN{} items. By point estimate the margins of the two strongest VLMs are larger on the other plots (\PlotQAOtherTopDeltas) than on the horizontal bar charts, but the horizontal bar charts, \PlotQAHorizontalItemShare{} of the items, still account for \PlotQAHorizontalShareTop{} of the pooled margins of these two VLMs. Every VLM scores separably lower on the horizontal bar charts than on the other plots (unpaired intervals), by \PlotQAOrientationGapMin{} to \PlotQAOrientationGapMax~points; DePlot's difference of \PlotQAOrientationGapDePlot~points is not separable (unpaired interval \PlotQAOrientationGapCIDePlot).

The second is the scoring convention. Under whole-table scoring \PlotQAWholeMovedName{} is no longer separable from DePlot (\PlotQAWholeMovedDelta). Our scores also count a value that occurs several times in a series only once, which penalises a reply that rounds nearby values to the same number: that number matches only one of them. Counting every repeated value separately removes that penalty, but a reply that omits a repeated value then loses a match for each omission. With repeats counted every score rises (by \PlotQARepeatsGainDePlot~points for DePlot and by up to \PlotQARepeatsGainMax~points for \PlotQARepeatsGainMaxName). Both of the strongest VLMs are then separably above DePlot (\PlotQARepeatsTopDeltas), and \PlotQAWholeMovedName, separably below DePlot in Table~\ref{tab:plotqa}, is not separable from it (\PlotQARepeatsMovedDelta). All three scorings (best-series, whole-table and best-series with repeats counted) order the seven systems in the same way, and \PlotQAAlwaysBelowList{} are separably below DePlot under each.

The third is the magnitude of the values, a split we made after seeing the results. On the \PlotQASmallN{} items whose median absolute value is below $10^{\PlotQALargeExponent}$ the two strongest VLMs score \PlotQASmallTopDeltas{} points against DePlot; on the other \PlotQALargeN{} they lead it by \PlotQALargeTopDeltas{} points. DePlot's deficit against the two strongest VLMs is thus concentrated in these \PlotQALargeN{} charts, which, by point estimate, account for \PlotQALargeShareTop{} of the two pooled margins. Of these charts, \PlotQALargeHorizontalN{} are horizontal bar charts and \PlotQALargeOtherN{} are other plots, about the same mix as among all \PlotQAN{} items, so this is not the orientation split again. At every power of ten from $10^{\PlotQAThresholdLowExponent}$ to $10^{\PlotQAThresholdHighExponent}$ the two strongest VLMs lead DePlot separably on the items at or above it and do not lead it separably on those below it.

Two further observations concern DePlot's replies. In \PlotQADePlotShortReplies{} of the \PlotQAN{} replies, DePlot's table has one row fewer than the annotated series has values, which never happens in the replies of the two strongest VLMs. On those \PlotQADePlotShortReplies{} items DePlot scores \PlotQADePlotShortScore\%, against \PlotQADePlotRestScore\% on the rest; we do not know whether this is the model or our run settings. And among the annotated values that a reply reproduces within the tolerance, DePlot's are the closest: their median relative error is \PlotQAMedErrDePlot\%, against \PlotQAMedErrVLMMin--\PlotQAMedErrVLMMax\% for the VLMs.

Across best-series, whole-table and repeats-counted scoring the pooled margins of the two strongest VLMs stay between \PlotQATopMarginMin{} and \PlotQATopMarginMax~points, and the \PlotQAAlwaysBelowCount{} weakest VLMs stay separably below DePlot. Neither statement holds in every subset: the two strongest lead by much more on the \PlotQALargeN{} items with very large values, and \PlotQAOtherLevelWeakName{} is level with DePlot on the \PlotQAOtherN{} other plots (\PlotQAOtherLevelWeakDelta).

\FloatBarrier
\subsection{Sensitivity to the tolerance}
\label{sec:tolerance}

Table~\ref{tab:tolerance} repeats both comparisons at tolerances of 1, 2, 5 and 10\%. Statements in this section are by point estimate unless they say separable or give an interval. On ChartX the picture is mixed. The margin over DePlot is larger at 1\% than at \Tolerance\% for \TolChartXMarginGrows{} of the \ChartXModelCount{} VLMs and smaller for \TolChartXMarginShrinksList. \ChartXLowestVLMName{} is not separable from DePlot at 1\% (\TolChartXOneLowestDelta) or at 2\% (\TolChartXTwoLowestDelta); the other \TolChartXAboveOne{} VLMs lead DePlot separably at every tolerance, \TolChartXOneSecondLowestName{} only just at 1\% (\TolChartXOneSecondLowestDelta).

On PlotQA the tolerance matters more, and in DePlot's favour: every VLM loses more than DePlot as the tolerance tightens, and for \TolPlotQADropDoubled{} of the \PlotQAModelCount{} the fall from 10\% to 1\% is more than twice DePlot's. At 1\% \TolPlotQATieName{} ties DePlot (\TolPlotQAOneDePlot\% each) and \TolPlotQAOneSecondName{} is separably below it (\TolPlotQAOneSecondDelta); at 10\% both are separably above it (\TolPlotQATenTopDeltas). This is the other side of the median relative error noted in Section~\ref{sec:plotqa}: among the values that a reply reproduces within \Tolerance\%, DePlot's are closer to the annotated values than those of any VLM, and we did not test why. The standing of the two strongest VLMs against DePlot on PlotQA therefore depends on the \Tolerance\% tolerance, which was chosen after results had been seen, while the other \PlotQABelowCount{} are below DePlot at every tolerance, all separably except \TolPlotQATenLevelName{} at 10\% (\TolPlotQATenLevelDelta). The ChartX comparison does not depend on it, except for \ChartXLowestVLMName, and for \TolChartXOneSecondLowestName{} at 1\%, where its lead is borderline.

\begin{table}[ht]
  \centering
  \caption{Scores (\%) of every system at relative tolerances of 1, 2, 5 and 10\%: numeric F1 on the ChartX validation split and best-series numeric F1 on the PlotQA subset. The two blocks use different benchmarks and variants of the metric and are not comparable with each other. The \Tolerance\% columns repeat Tables~\ref{tab:chartx} and~\ref{tab:plotqa}. Rows follow Table~\ref{tab:chartx}. A dash marks a model that was not run on PlotQA.}
  \label{tab:tolerance}
  \small
  \begin{tabular}{lrrrr@{\hspace{2em}}rrrr}
    \toprule
          & \multicolumn{4}{c}{ChartX} & \multicolumn{4}{c}{PlotQA} \\
    \cmidrule(lr){2-5}\cmidrule(lr){6-9}
    Model & 1\% & 2\% & 5\% & 10\% & 1\% & 2\% & 5\% & 10\% \\
    \midrule
    Gemini 3 Flash        & \TolChartXOneGeminiFlash     & \TolChartXTwoGeminiFlash     & \TolChartXFiveGeminiFlash     & \TolChartXTenGeminiFlash     & \TolPlotQAOneGeminiFlash     & \TolPlotQATwoGeminiFlash     & \TolPlotQAFiveGeminiFlash     & \TolPlotQATenGeminiFlash \\
    Gemini 3.1 Flash-Lite & \TolChartXOneGeminiFlashLite & \TolChartXTwoGeminiFlashLite & \TolChartXFiveGeminiFlashLite & \TolChartXTenGeminiFlashLite & \TolPlotQAOneGeminiFlashLite & \TolPlotQATwoGeminiFlashLite & \TolPlotQAFiveGeminiFlashLite & \TolPlotQATenGeminiFlashLite \\
    GPT-5.4 mini          & \TolChartXOneGptMini         & \TolChartXTwoGptMini         & \TolChartXFiveGptMini         & \TolChartXTenGptMini         & \TolPlotQAOneGptMini         & \TolPlotQATwoGptMini         & \TolPlotQAFiveGptMini         & \TolPlotQATenGptMini \\
    Claude Sonnet 4.6     & \TolChartXOneSonnet          & \TolChartXTwoSonnet          & \TolChartXFiveSonnet          & \TolChartXTenSonnet          & \TolPlotQAOneSonnet          & \TolPlotQATwoSonnet          & \TolPlotQAFiveSonnet          & \TolPlotQATenSonnet \\
    GPT-5.4 nano          & \TolChartXOneGptNano         & \TolChartXTwoGptNano         & \TolChartXFiveGptNano         & \TolChartXTenGptNano         & \TolPlotQAOneGptNano         & \TolPlotQATwoGptNano         & \TolPlotQAFiveGptNano         & \TolPlotQATenGptNano \\
    Claude Haiku 4.5      & \TolChartXOneHaiku           & \TolChartXTwoHaiku           & \TolChartXFiveHaiku           & \TolChartXTenHaiku           & \TolPlotQAOneHaiku           & \TolPlotQATwoHaiku           & \TolPlotQAFiveHaiku           & \TolPlotQATenHaiku \\
    Mistral Medium 3.5    & \TolChartXOneMistralMedium   & \TolChartXTwoMistralMedium   & \TolChartXFiveMistralMedium   & \TolChartXTenMistralMedium   & -- & -- & -- & -- \\
    Mistral Small 4       & \TolChartXOneMistralSmall    & \TolChartXTwoMistralSmall    & \TolChartXFiveMistralSmall    & \TolChartXTenMistralSmall    & -- & -- & -- & -- \\
    Mistral Large 3       & \TolChartXOneMistralLarge    & \TolChartXTwoMistralLarge    & \TolChartXFiveMistralLarge    & \TolChartXTenMistralLarge    & -- & -- & -- & -- \\
    \midrule
    DePlot                & \TolChartXOneDePlot          & \TolChartXTwoDePlot          & \TolChartXFiveDePlot          & \TolChartXTenDePlot          & \TolPlotQAOneDePlot          & \TolPlotQATwoDePlot          & \TolPlotQAFiveDePlot          & \TolPlotQATenDePlot \\
    \bottomrule
  \end{tabular}
\end{table}

\subsection{Prompt sensitivity}
\label{sec:prompt}

Every result in Tables~\ref{tab:chartx} to~\ref{tab:tolerance} and in both figures uses the simple two-sentence prompt; a longer prompt was used only in the Claude development-split runs of Section~\ref{sec:chartx} and in the two PlotQA runs compared here. Version 1 reported that a detailed prompt improved PlotQA scores by 1 to 3 points, which rested on the mis-scored results (Section~\ref{sec:plotqa}). Scored against the plotted values, the released detailed-prompt runs of Claude Sonnet 4.6 and Claude Haiku 4.5 reach \PromptDetailedSonnet\% and \PromptDetailedHaiku\% and differ from the simple-prompt runs by \PromptDeltaSonnet~\PromptDeltaCISonnet{} and \PromptDeltaHaiku~\PromptDeltaCIHaiku~points. Neither difference is separable from zero, and the upper ends of the intervals are \PromptDeltaHighSonnet~points for Claude Sonnet 4.6 and \PromptDeltaHighHaiku~points for Claude Haiku 4.5, so the gain of 1 to 3 points that version 1 reported is not supported. This rests on one run per prompt, made on different dates. Which run used which prompt is known only from the history of the development repository, which is not part of the release, and the comparison says nothing about ChartX or about the application's own, much longer, structured prompt.

\FloatBarrier

\section{Discussion}
\label{sec:discussion}

The two benchmarks point in different directions. On the \ChartXTypeCount{} ChartX chart types all \ChartXModelCount{} VLMs outperform DePlot in aggregate, across all \ProviderCount{} providers. The lead is largest by far on box plots; on the other \ChartXNonBoxTypeCount{} chart types the \ChartXTopCount{} strongest VLMs still lead DePlot in every cell, by \ChartXTopCellMin{} to \ChartXTopCellMax~points, and, pooled over those types, the \ChartXTrailingModels{} weakest do not lead it. On PlotQA, whose training split was part of DePlot's training data, the two strongest VLMs are level with DePlot or slightly above it and the other \PlotQABelowCount{} are behind it, and a stricter tolerance moves the comparison there in DePlot's favour. DePlot is thus competitive with the strongest of these VLMs on the benchmark that comes from its own training distribution, and behind the \ChartXTopCount{} strongest on the bar and line charts of ChartX, which belong to chart families in its training data but were drawn by another generator. We ran no experiment that isolates the causes. Three things may contribute:

\begin{enumerate}
  \item \textbf{Chart-type coverage}: DePlot's plot-to-table training data is an equal mix of synthetic charts, ChartQA and PlotQA, and its authors describe line, dot, bar and pie charts~\cite{liu2023deplot}. The ChartQA charts and the synthetic charts are bar, line and pie charts~\cite{masry2022chartqa,liu2023matcha}, and PlotQA adds dot-line plots~\cite{methani2020plotqa}; box plots are not among them. DePlot's box-plot replies hold one value per box (Section~\ref{sec:chartx}), as if it read a box plot as a bar chart: a failure of output format on a chart type outside its documented training data. Histograms are not among the documented types either, yet DePlot scores \ChartXHistogramDePlot\% on them. About half of the ChartX histograms have category names rather than numeric bins: in only \ChartXHistogramNumeric{} of the \ChartXPerType{} does every row label contain a number.
  \item \textbf{Familiarity with the benchmark}: DePlot's PlotQA score falls less than that of any VLM as the tolerance tightens (Table~\ref{tab:tolerance}), and the values it reproduces are the closest to the annotated values (Section~\ref{sec:plotqa}). Two things could produce this and we cannot separate them. DePlot was trained on PlotQA's training split, drawn by the same generator. And PlotQA's 224,377 plots were generated from combinations of 841 indicator variables and 160 entities and then divided into training, validation and test splits~\cite{methani2020plotqa}; that paper does not say whether the splits share data series. Within our \PlotQASubsetN{} test entries, \PlotQARepeatedSeries{} pairs share an identical series of varying values, so the same numbers do recur across charts, and the numbers behind a test chart may also occur in a training chart.
  \item \textbf{Printed values}: every system, DePlot included, scores higher on the charts that print their values as data labels than on the other charts of the same family, by \ChartXLabelGainMin{} to \ChartXLabelGainMax~points. These are different charts; the \ChartXLabelComparisons{} system-by-family comparisons share the same charts and are not independent; and in \ChartXLabelSpanZero{} of them the unpaired 95\% interval includes zero, so the data do not show that the labels are the cause.
\end{enumerate}

Model scale is a further candidate that we cannot examine, because the sizes and training data of most of the VLMs are not disclosed.

Neither benchmark resembles a published biomedical figure, so we do not rank the two for the intended use. Charts that do not print their values are the case a figure-extraction tool exists for. On the \ChartXUnlabelledTypeCount{} ChartX chart types without data labels, Claude Haiku 4.5, the one model the application offers that we benchmarked, scores \ChartXHaikuUnlabelledMin--\ChartXHaikuUnlabelledMax\% numeric F1. On the PlotQA subset, whose plots carry no data labels either as far as the PlotQA paper shows, it scores \PlotQABestHaiku\% best-series numeric F1, a more lenient score. The two benchmarks differ in generator, annotation and metric, so these numbers are not comparable. The application's own prompt asks for the median and quartiles of each box but not for the whiskers or outliers, which are part of the ChartX box-plot ground truth, so the box-plot result does not carry over directly either.

\section{Limitations}
\label{sec:limitations}

What these results establish is limited. On synthetic charts, some general-purpose VLMs recover plotted numbers as reliably as DePlot or more so, most clearly on box plots. Nothing here shows how often PlotPick returns the right value for the right group on a published figure.

First, what was evaluated is the reading of numbers from chart images by VLMs of the kind the application calls, only one of which it offers. The application's figure detection, its structured prompt, the fields it returns beyond the values themselves (error bars, error-bar type, group size, group and timepoint labels) and its default model were not evaluated. The one model it offers that we did benchmark, Claude Haiku 4.5, scores below Claude Sonnet 4.6 on both benchmarks (\ChartXFoneHaiku\% against \ChartXFoneSonnet\% numeric F1 on the \ChartXTypeCount{} ChartX chart types, \PlotQABestHaiku\% against \PlotQABestSonnet\% best-series numeric F1 on the PlotQA subset).

Second, both benchmarks consist of synthetic charts. We have not established that this accuracy transfers to figures in the biomedical literature, which is the setting that motivates the tool. We did build an exploratory set of figure--table pairs from PubMed Central articles. We ran two Claude models (Haiku and Sonnet) on it, and Claude Opus on \PMCPairsOpus{} of its pairs (not reported here): each model was given an empty copy of the figure's companion table, with its row labels, and asked to fill the cells with numbers that the figure prints as text. That run is not an accuracy measurement in either direction: the tables hold a median of \PMCMedianCells{} numeric cells, far more than a figure shows, and the task excluded reading values from axis scales, so its mean recall of \PMCRecallHaiku\% (Haiku, \PMCPairsHaiku{} pairs) and \PMCRecallSonnet\% (Sonnet, \PMCPairsSonnet{} pairs) measures how much of a table is printed in its figure rather than how accurately a figure is read. We draw no conclusion from it. A properly scoped evaluation on real trial figures, with ground truth restricted to values actually plotted, remains future work and is the comparison we regard as most important.

Third, our metrics compare unlabelled sets of numbers at a \Tolerance\% tolerance. A value read correctly but assigned to the wrong group or timepoint is scored as correct, so is a value up to \Tolerance\% from the truth, and so, on ChartX, is a number that appears only in a label, such as a year or the 1 of ``Q1'' (Section~\ref{sec:chartx}). For meta-analysis, where the pairing of value to arm is the point, the first is the error mode that can matter most and we do not measure it. The second matters for pooling: two arm means of \SmdExampleMeans{} with a standard deviation of \SmdExampleSD{} give a standardised mean difference of \SmdExampleTrue, and readings \Tolerance\% off in opposite directions can at worst turn that into anything from \SmdExampleLow{} to \SmdExampleHigh, even with the standard deviation read exactly.

Fourth, the PlotQA items are not a sample we designed. They are the first \PlotQASubsetN{} entries of a converted test split, \PlotQASubsetHorizontal{} of them horizontal bar charts; the \PlotQAN{} scored ones were further selected by the error described in Section~\ref{sec:plotqa}; the ground truth holds one series per chart; and the best-series score is optimistic. We validated the corrected scoring against the annotations and the replies, not by viewing the chart images.

Fifth, DePlot is one chart-specific model, published in 2023. Newer chart-specific models, among them TinyChart~\cite{zhang2024tinychart} and ChartVLM from the ChartX authors~\cite{chartx2024}, were not run, so the comparison is with one earlier specialist and not with the state of the art.

Sixth, we report no comparison against the interactive digitisers used in systematic reviews, such as WebPlotDigitizer, with or without its AI Assist, or metaDigitise, on the same figures. WebPlotDigitizer has been evaluated for validity on single-case-design graphs~\cite{drevon2017}, and agreement within our tolerance is not evidence of equivalent precision.

Seventh, both benchmarks are public and the training data of the VLMs are not disclosed, so the benchmark charts may have been in that training data. DePlot cannot have seen ChartX, which was published after it, but it was trained on PlotQA's training split (Section~\ref{sec:discussion}). Each system was run once per benchmark and prompt, and the intervals reflect the sample of items, not the variation between repeated runs of a model. Two identifiers are previews, one of which can no longer be called. The PlotQA comparison covers \PlotQAModelCount{} of the \ChartXModelCount{} models. The chart types, the tolerance, the scoring and the magnitude split were chosen after results had been seen; the VLMs ran at provider defaults that differ between models, with a reply cap for the Claude and OpenAI models; DePlot received the original image and the VLMs at most \ImageMaxSide{} pixels; and the ChartX validation items were not audited for ground-truth errors.

Every extracted value must be checked against its source figure by a person, and we did not measure whether the tool saves time. The per-figure confidence score and the per-row uncertainty flags are self-reported by the model and untested, so neither says which values can be skipped.

\section{Conclusion}

On \ChartXTypeCount{} ChartX chart types, the \ChartXModelCount{} general-purpose VLMs we tested, given a two-sentence prompt and no fine-tuning by us, lead the dedicated model DePlot in aggregate, by \ChartXGapLowestVLM{} to \ChartXGapHighestVLM~points of numeric F1, with much of that lead on box plots; the lowest of these margins, that of \ChartXLowestVLMName, is not separable from DePlot at a tolerance of 1\% or 2\%, or when the digits in labels are left out. On a PlotQA subset, by a lenient best-series variant of the metric, the two strongest VLMs are level with DePlot or slightly above it at the \Tolerance\% tolerance, though at 1\% and on the \PlotQASmallN{} items without very large values \TolPlotQAOneSecondName{} is below DePlot (Sections~\ref{sec:plotqa} and~\ref{sec:tolerance}), and DePlot leads the other \PlotQABelowCount{} by \PlotQABelowMin{} to \PlotQABelowMax~points. General-purpose VLMs are therefore not uniformly better than DePlot: it is level with the two strongest of them, or slightly behind them, on PlotQA, the benchmark drawn from its own training distribution, and behind all of them in aggregate on the \ChartXTypeCount{} ChartX chart types, most of all on box plots, a chart type outside its documented training data. These results concern DePlot only.

PlotPick wraps VLMs in an application with figure detection and batch upload. Its own prompt, its default model and its figure detection were not evaluated, and the one benchmarked model that it offers, Claude Haiku 4.5, scores \ChartXFoneHaiku\% on the \ChartXTypeCount{} ChartX chart types, above DePlot's \ChartXFoneDePlot\%, and \PlotQABestHaiku\% best-series numeric F1 on the PlotQA subset, below DePlot's \PlotQABestDePlot\%; the two scores are not comparable with each other. PlotPick is intended to reduce the effort of extracting data from figures for systematic reviews and meta-analyses in the biomedical sciences; we did not measure time savings, and every extracted value has to be checked against its figure.

\section{Changes from version 1}
\label{sec:changes}

Version 1 of this preprint stated that all six VLMs it reported outperformed DePlot on both benchmarks, that general-purpose VLMs outperformed dedicated chart-to-table models on every chart type tested, and that PlotPick outperformed all published dedicated models on PlotQA. Those statements are withdrawn. Of the dedicated models, only DePlot was run; in this version all \ChartXModelCount{} VLMs lead it in aggregate on the \ChartXTypeCount{} ChartX chart types, and on PlotQA two of \PlotQAModelCount{} are level with it or slightly above it. The evidence of version 1 was wrong in the following ways.

\paragraph{PlotQA.} On \PlotQAHorizontalN{} of the \PlotQAN{} items every score was computed against the wrong axis (Section~\ref{sec:plotqa}), so the 86--99\% that version 1 reported for the VLMs mostly measured whether a reply contained a column of years. Version 1 set those scores beside DePlot's published 94.2\%, which is an RMS F1 on the full PlotQA test set and not the metric applied to the VLMs, and it did not report our own run of DePlot on the same items. Even on its own numbers the claim did not hold for GPT-5.4 nano (86.3\% against 94.2\%). Scored against the plotted values, DePlot is ahead of \PlotQABelowCount{} of the \PlotQAModelCount{} VLMs.

\paragraph{ChartX.} The DePlot figure of 70.5\%, which version 1's Introduction gave as DePlot's score on ChartX with a citation of the ChartX paper, was our own F1 on \DevSixN{} development-split items. Version 1's table printed it with $N{=}\DevSixN$ in a column of recall, beside the VLMs' recall on 300 validation-split items, under a caption that named the validation split. Figures 1 and 2 of version 1 showed the development split without saying so (Figure 1 with the VLMs' recall and DePlot's F1 on one axis), while the VLM rows of its table showed the validation split: in the figures, DePlot was scored on \DevDePlotPerType{} items per chart type and the VLMs on \DevPerTypeMin{} or \DevPerTypeMax, and the two Claude models had been given a longer prompt although the text said that all experiments used the same simple prompt. The box-plot comparison in the abstract of version 1 (DePlot 24\% against 83--97\% for the VLMs) took the 24\% from those figures, where it is DePlot's score on \DevDePlotPerType{} development-split box plots, but the range given for the VLMs matches neither split: the six VLMs score \DevBoxFoneMin--\DevBoxFoneMax\% numeric F1 on the box plots of the development split, as Figure 2 of version 1 shows, and \ValBoxSixFoneMin--\ValBoxSixFoneMax\% on those of the validation split. Version 1's claim of a lead on every chart type, made in its Introduction and in the caption of its Figure 2, also failed in that figure, where Claude Haiku 4.5 scores 91 and DePlot 93 on plain line charts. In this version DePlot is run on the same \ChartXNDePlot{} items and scored by the same code, without its title line (Section~\ref{sec:metrics}). The conclusion that the VLMs lead in aggregate stands, with a smaller margin and a clearer account of where it comes from.

\paragraph{Withdrawn statements.} That a detailed prompt improved scores by 1 to 3 points rested on the mis-scored PlotQA results (Section~\ref{sec:prompt}). That image upscaling improved accuracy by about 3 points has no retained measurement behind it. That charts with data labels reached 98\% or more for every VLM is not what the results show: the six VLMs of version 1 score \ChartXLabelledSixMin--\ChartXLabelledSixMax\% on the two chart types with data labels (Figure~\ref{fig:heatmap} of this version). That dedicated models had never seen box plots or histograms, that stacked and grouped bar charts are systematically harder and that GPT-5.4 nano misreads magnitudes were not measured, and are removed; this version says only that box plots and histograms are not among the chart types documented for DePlot's training data (Section~\ref{sec:discussion}). That DePlot and TinyChart generalise poorly to the chart types of the biomedical literature was not tested: TinyChart was not run, and DePlot was run on no biomedical figure.

\paragraph{The application.} Version 1 said that the application sends the benchmark's simple prompt and that any VLM accepting image input could be substituted. It sent, and sends, a longer structured prompt that was not benchmarked, and it has only ever had an Anthropic backend. Version 1 also described a single-file application with editable tables and a confidence score per row; the application had more than one module, its tables are read-only and the confidence score is per figure (Section~\ref{sec:design}). Its default model was Claude Haiku 4.5 when version 1 was posted and is now Claude Sonnet 5.5, and it now reads PDFs with pypdfium2 instead of PyMuPDF.

\paragraph{Other corrections.} Version 1 said its results were computed after correcting ground-truth errors in ChartX. That was misleading: the corrected charts are all in the development split and none is among the \DevDePlotN{} on which DePlot was run there, so the corrections could bear only on the VLM results that version 1 took from that split (its Figures 1 and 2 and the numbers quoted from them) and, in this version, only on the development-split results quoted in Section~\ref{sec:chartx}, not on any table or figure. Version 1 said the benchmark scripts and results were in the application repository; they were not, and they are now released in a repository of their own. Version 1 cited a study of Plot Digitizer as evidence for WebPlotDigitizer and said that existing digitisers ``do not produce structured, labelled output suitable for downstream meta-analysis''; both are corrected in the Introduction. It listed Claude Haiku 4.5 with 300 items where the run has \ChartXNHaiku. Its AI usage disclosure understated the use of AI: it described Claude Code as assisting with the benchmark scripts, the data analysis and the drafting, and said that the application itself was developed by the author, whereas Claude Code wrote or revised most of the code of the application and of the benchmark; the disclosure below replaces it. The metric that version 1 called RMSF1 is renamed numeric F1 (Section~\ref{sec:metrics}). Mistral models had been run under moving aliases before version 1 was posted, in \AliasRunCount{} runs on the ChartX validation split, \AliasDevRunCount{} on its development split and \AliasPlotQARunCount{} on PlotQA, and were left out of it; Sections~\ref{sec:chartx} and~\ref{sec:plotqa} say how those on the ChartX validation split and on PlotQA score. This version reports \MistralModelCount{} Mistral models run under dated identifiers, on ChartX only, and every number of this version that describes our own results is generated from the released per-item results; the figures quoted from version 1 are copied from that version, and the ranges given for its six VLMs are recomputed.

\section*{Data and code availability}

PlotPick is open source under the MIT licence: \url{https://github.com/tommycarstensen/plotpick}

The benchmark runners, the scoring and bootstrap scripts, the per-item result files and the generator of every number that describes our own results are released under the MIT licence: \\
\url{https://github.com/tommycarstensen/plotpick-validation} \\
The benchmark annotations redistributed there, and the ground-truth values embedded in the result files, keep their own licences: Apache-2.0 for ChartX, as stated on its dataset card, and CC-BY-4.0 for PlotQA; the PubMed Central values and metadata keep the licences of their articles. Tables~\ref{tab:chartx} to~\ref{tab:tolerance}, both figures and every interval quoted in the text can be regenerated from the stored per-item results without re-running any model or making any API call. The ChartX ground-truth corrections described above are included as an explicit list. The repository also holds the runs this paper does not tabulate (the development split, the other chart types, the moving-alias runs and the PubMed Central set), each described in its README.

\section*{AI usage disclosure}

Generative AI was used substantially in this work. The tool was Claude Code (Anthropic), using Claude Opus and Sonnet models. Claude Code wrote or revised most of the code in all parts of this project: the PlotPick application itself, including the Streamlit interface, the PDF figure-detection logic and the export functions; the benchmark and evaluation scripts; and the analysis and plotting code that produced the tables and figures reported here. It also drafted and revised most of the text of this manuscript. AI was also used to review this version: the errors described in Section~\ref{sec:changes} were found in several rounds of review by Claude models that were given the paper and the code, and the corrections were made with Claude Code. Claude Haiku 4.5 was additionally used as a component of the benchmark pipeline, to classify candidate figures and to screen candidate figure--table pairs while building the PubMed Central set described under Limitations, and Claude models are among the systems evaluated.

The author specified the research question, the software design and the choice of benchmarks, models and metrics, directed the work, and takes full responsibility for the accuracy of the software and of this paper.

\section*{Competing interests}

The author designed and ran this evaluation and is the developer of PlotPick, which is used in ongoing evidence synthesis at the author's research centre, the Copenhagen Research Centre for Biological and Precision Psychiatry. That use is not an evaluation, and no outcome of it is reported here. The application calls the API of Anthropic, one of the \ProviderCount{} providers whose models are benchmarked here.

\section*{Acknowledgements}

This work was supported by unrestricted grants from the Lundbeck Foundation (grant numbers R278-2018-1411 and R383-2022-285).

{\small
\bibliographystyle{unsrt}
\bibliography{main}
}

\end{document}